\documentclass{article}

\usepackage{PRIMEarxiv}

\usepackage[utf8]{inputenc} 
\usepackage[T1]{fontenc}    
\usepackage{hyperref}       
\usepackage{url}            
\usepackage{booktabs}       
\usepackage{amsfonts}       
\usepackage{nicefrac}       
\usepackage{microtype}      
\usepackage{lipsum}
\usepackage{fancyhdr}       
\usepackage{graphicx}       
\graphicspath{{media/}}     

\usepackage{amsmath}
\usepackage{multirow}
\usepackage{amssymb}
\usepackage{pifont}
\usepackage{xcolor}
\newcommand{\xmark}{\ding{55}}%

\title{Centered Masking for Language-Image Pre-Training

\author{
  Mingliang Liang, Martha Larson \\
  Radboud University \\
  Nijmegen\\
  \texttt{\{m.liang, m.larson\}@cs.ru.nl} \\
}
}

\begin{document}
\maketitle

\begin{abstract}
We introduce Gaussian masking for Language-Image Pre-Training (GLIP) a novel, straightforward, and effective technique for masking image patches during pre-training of a vision-language model.
GLIP builds on Fast Language-Image Pre-Training (FLIP), which randomly masks image patches while training a CLIP model.
GLIP replaces random masking with centered masking, which uses a Gaussian distribution and is inspired by the importance of image patches at the center of the image.
GLIP retains the same computational savings as FLIP, while improving performance across a range of downstream datasets and tasks, as demonstrated by our experimental results.
We show the benefits of GLIP to be easy to obtain, requiring no delicate tuning of the Gaussian, and also applicable to datasets containing images without an obvious center focus.

\end{abstract}

\section{Introduction}
The rise of Vision-Language Models trained on a large number of image text pairs has been led by Contrastive Language-Image Pretraining CLIP~\cite{CLIP2021radford}.
CLIP delivers high-quality visual representations, with strong performance on downstream tasks and impressive transferability.
However, the downside of CLIP, is the large amount of computational resources that it requires to train.
According to~\cite{CLIP2021radford}, pre-training CLIP on 400 million image-text pairs over 32 epochs required thousands of GPU days for completion.
The high computational cost of training Vision-Language models limits further increases in training data size.

Recent research has proposed to accelerate the training of CLIP, with surprising success being achieved by Fast Language-Image Pre-Training (FLIP)~\cite{li2023flip}. 
FLIP randomly masks image patches during CLIP training. 
According to~\cite{li2023flip}, if 50\% or 75\% of the patches in each training image are discarded by masking, FLIP can reduce the computation carried out during training by 2--4 times.
Interestingly, masking 50\% of an image does not compromise performance. 
In fact, it makes it possible to increase the global batch size given a processor's fixed processing power.
More samples per batch are known to be beneficial for contrastive learning ~\cite{he2022mae,He2020Momentum,li2023flip,CLIP2021radford}.

\begin{figure}[!t]
    \centering
    \begin{tabular}{cccccc}
    \includegraphics[width=0.15\textwidth,height=0.15\textwidth]{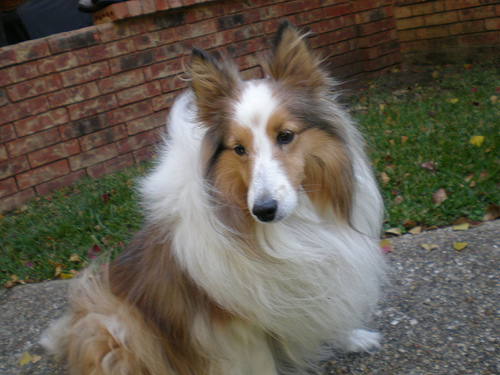}     
    & \includegraphics[width=0.15\textwidth,height=0.15\textwidth]{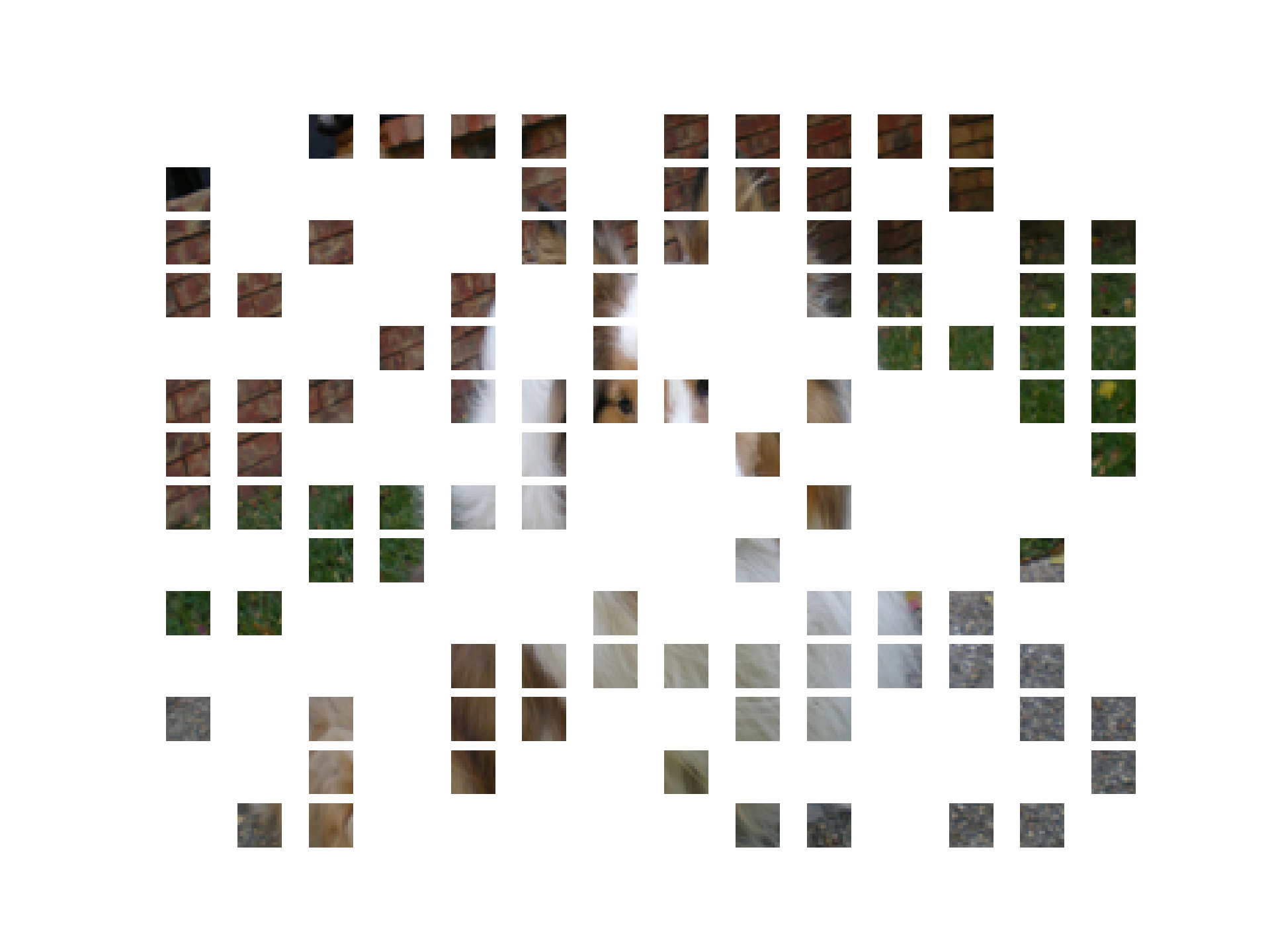}  
    & \includegraphics[width=0.15\textwidth,height=0.15\textwidth]{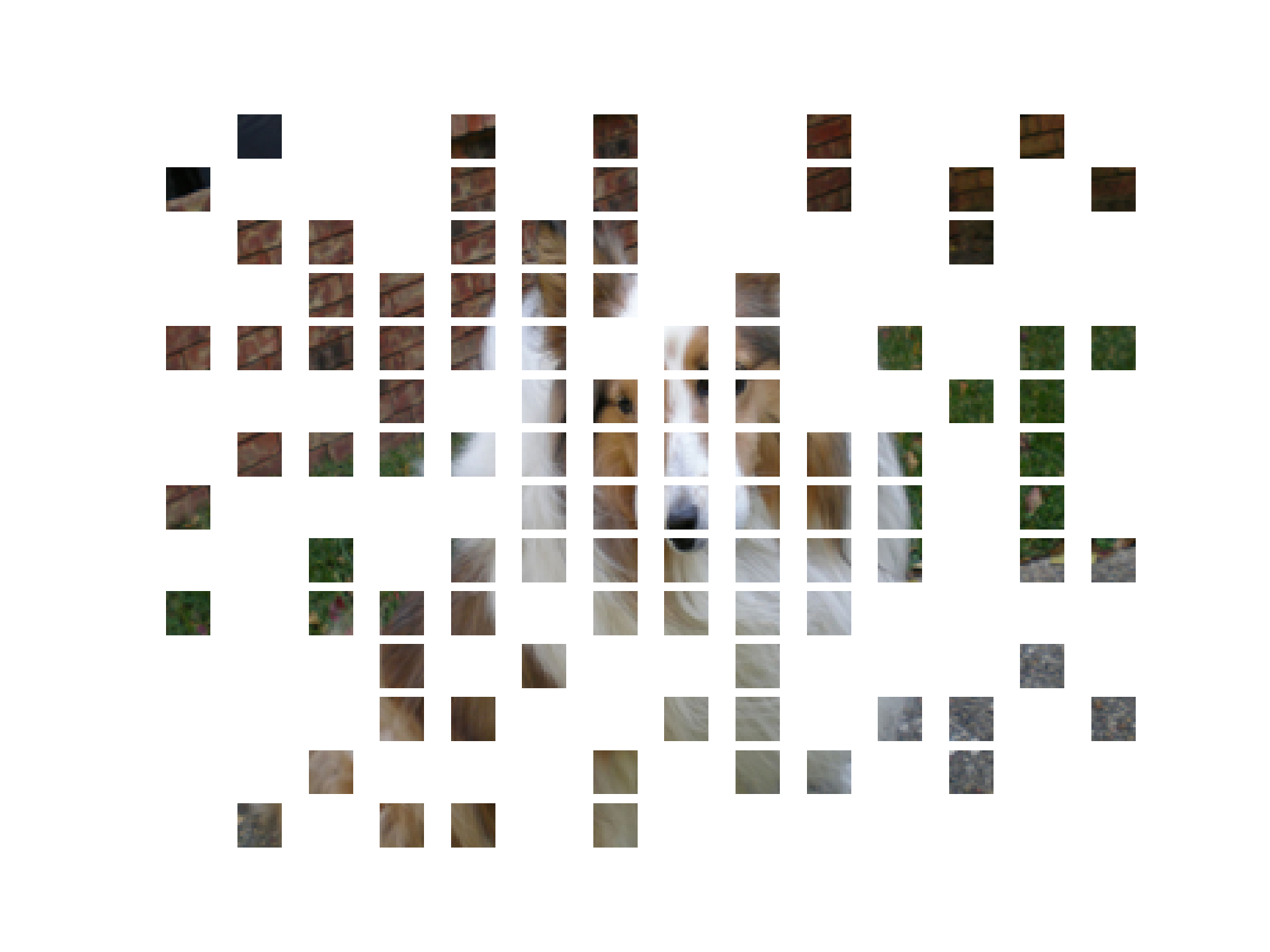}  
    \includegraphics[width=0.15\textwidth,height=0.15\textwidth]{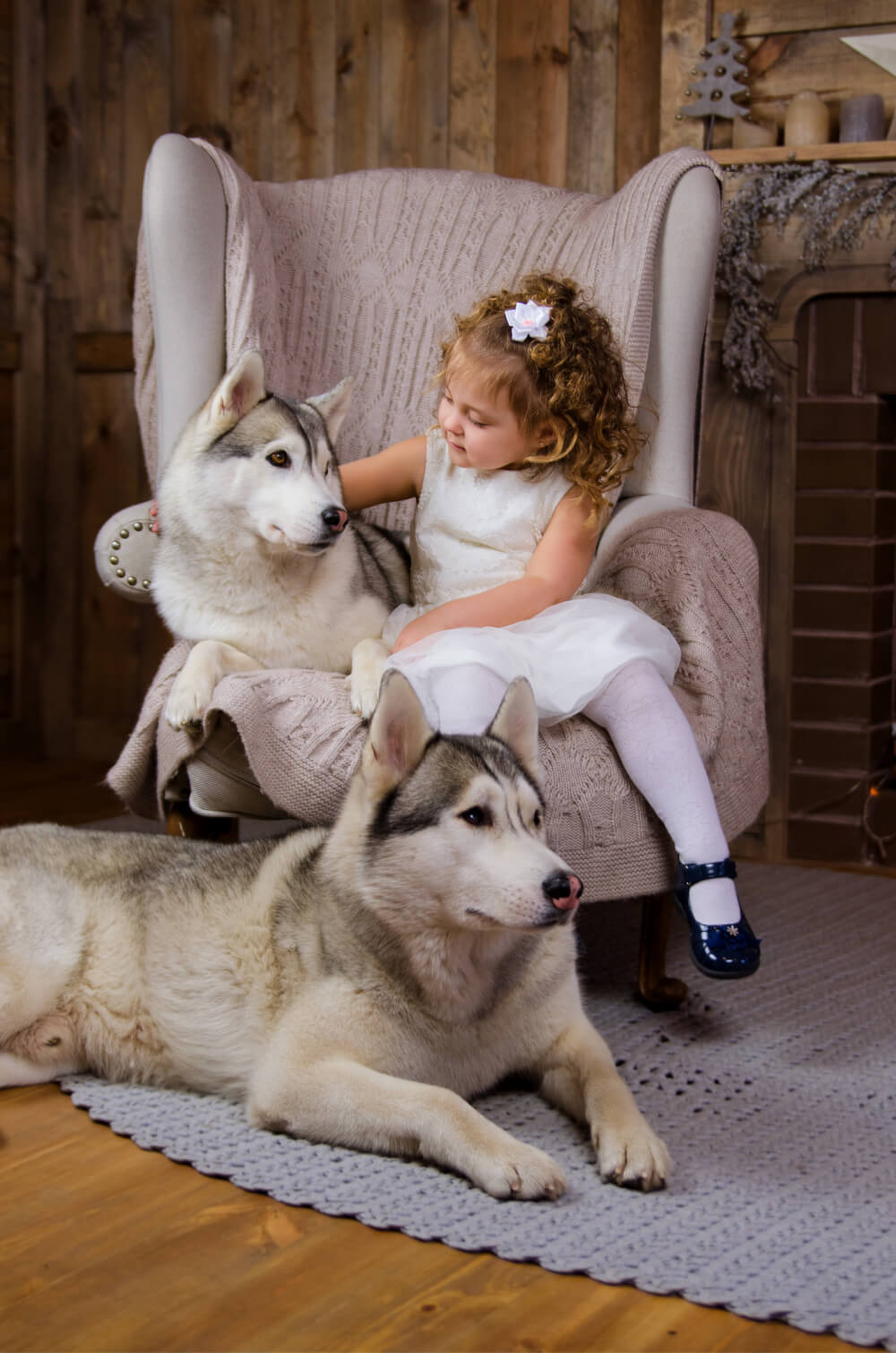} 
    \includegraphics[width=0.15\textwidth,height=0.15\textwidth]{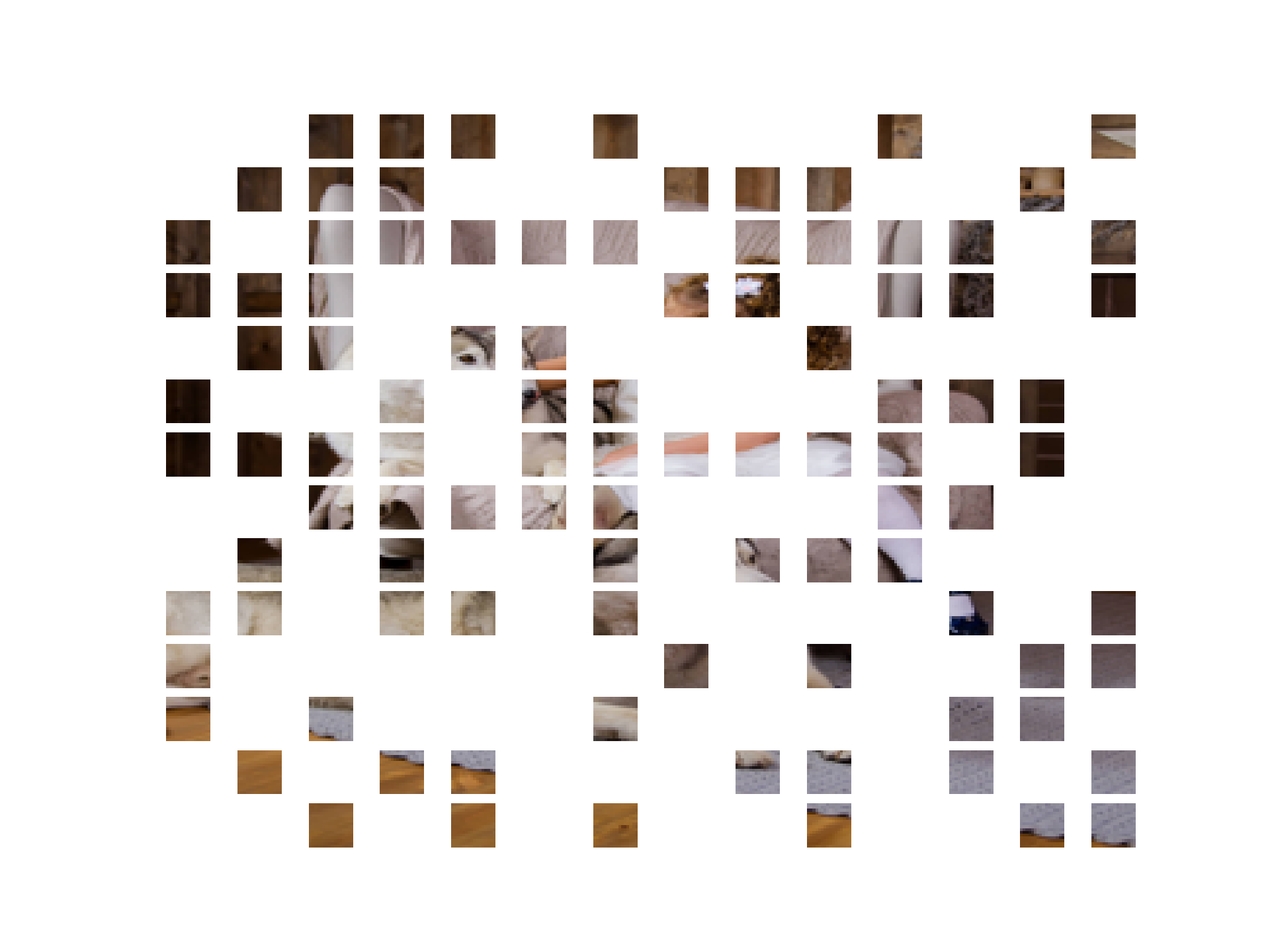} 
    \includegraphics[width=0.15\textwidth,height=0.15\textwidth]{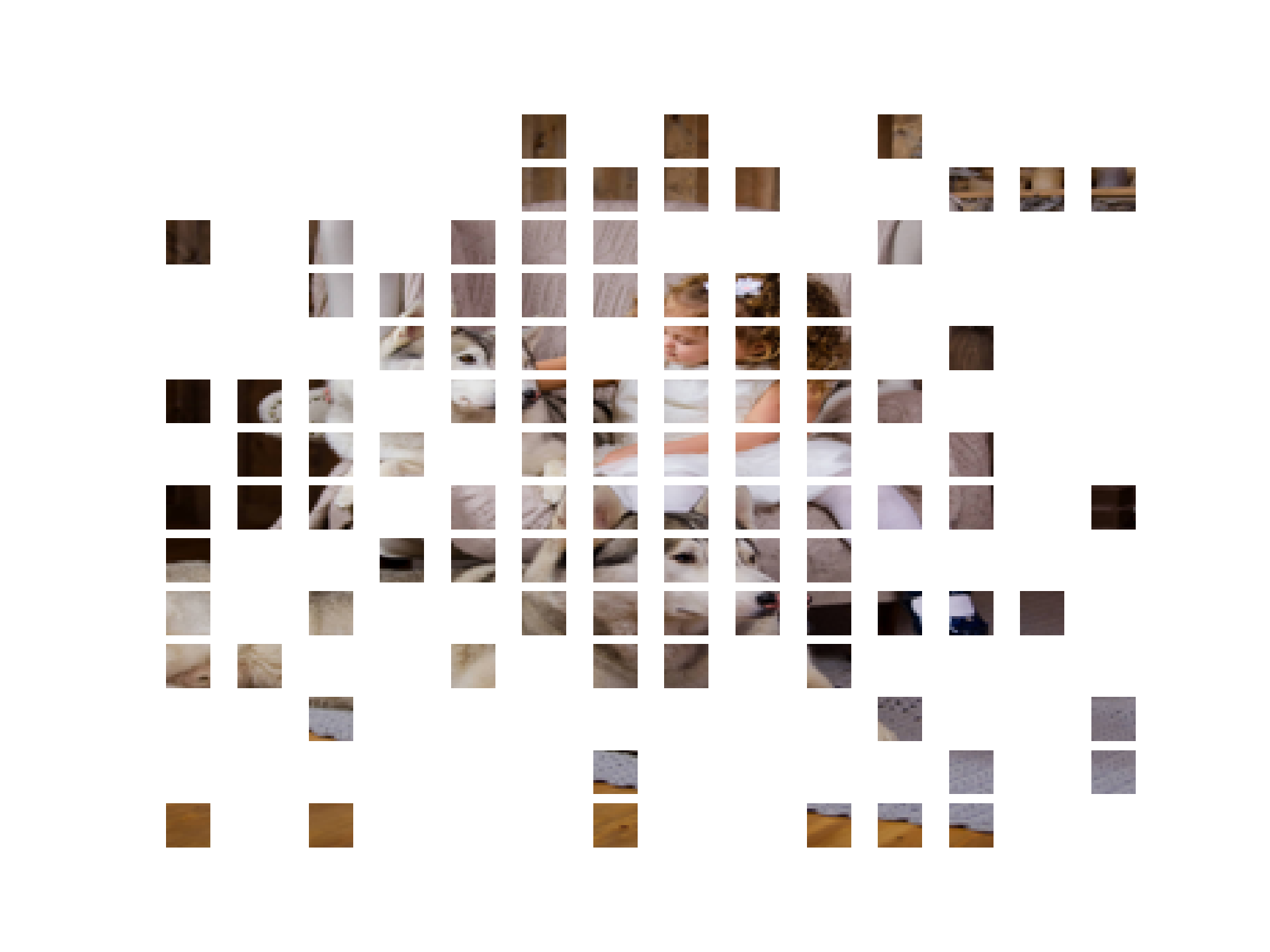} \\
     \includegraphics[width=0.15\textwidth,height=0.15\textwidth]{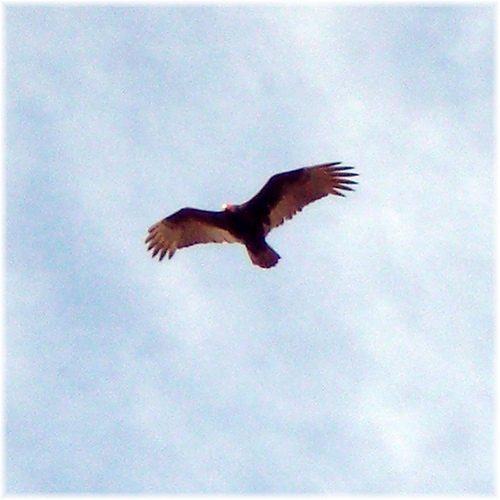} 
    & \includegraphics[width=0.15\textwidth,height=0.15\textwidth]{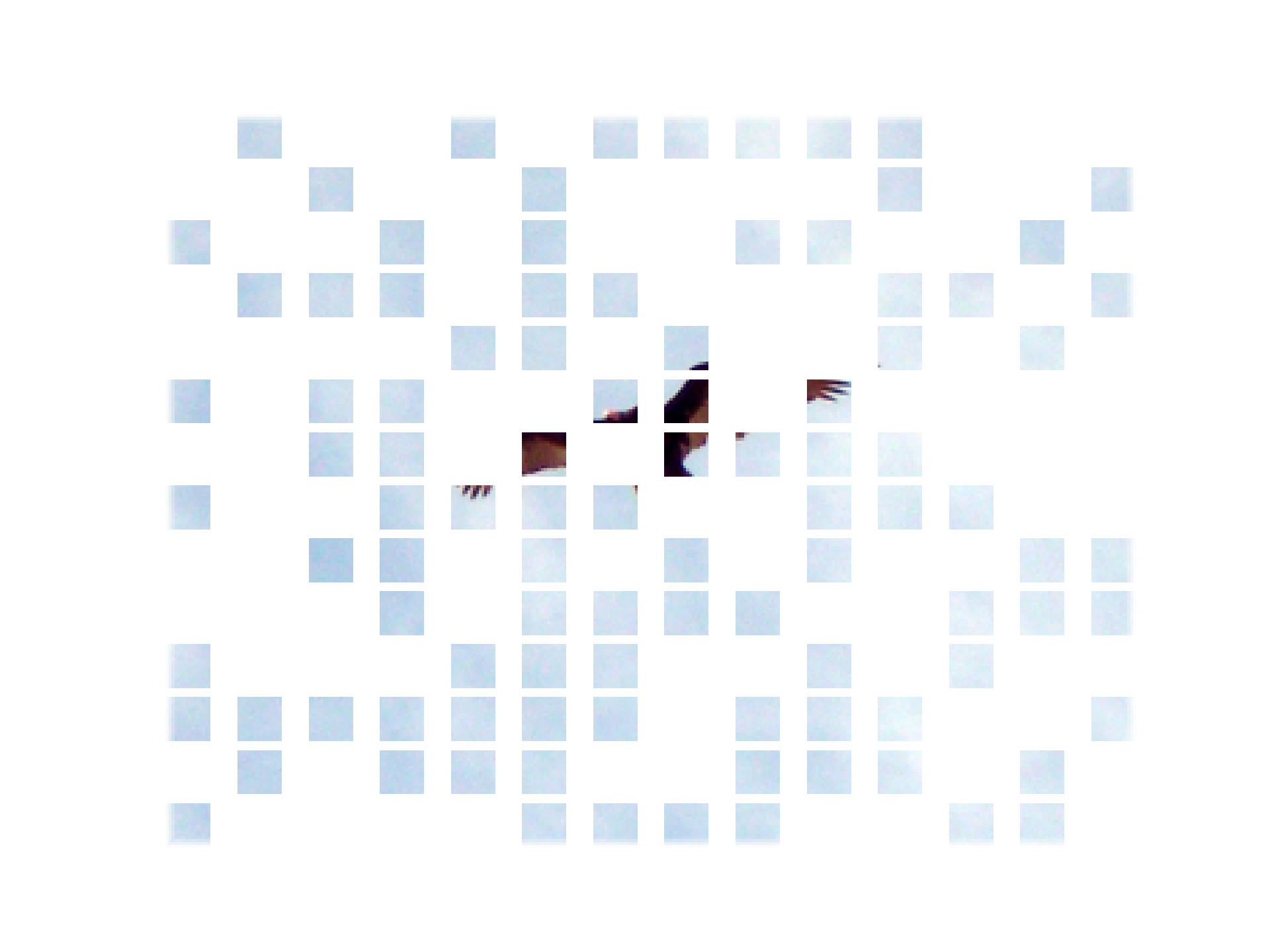}
    & \includegraphics[width=0.15\textwidth,height=0.15\textwidth]{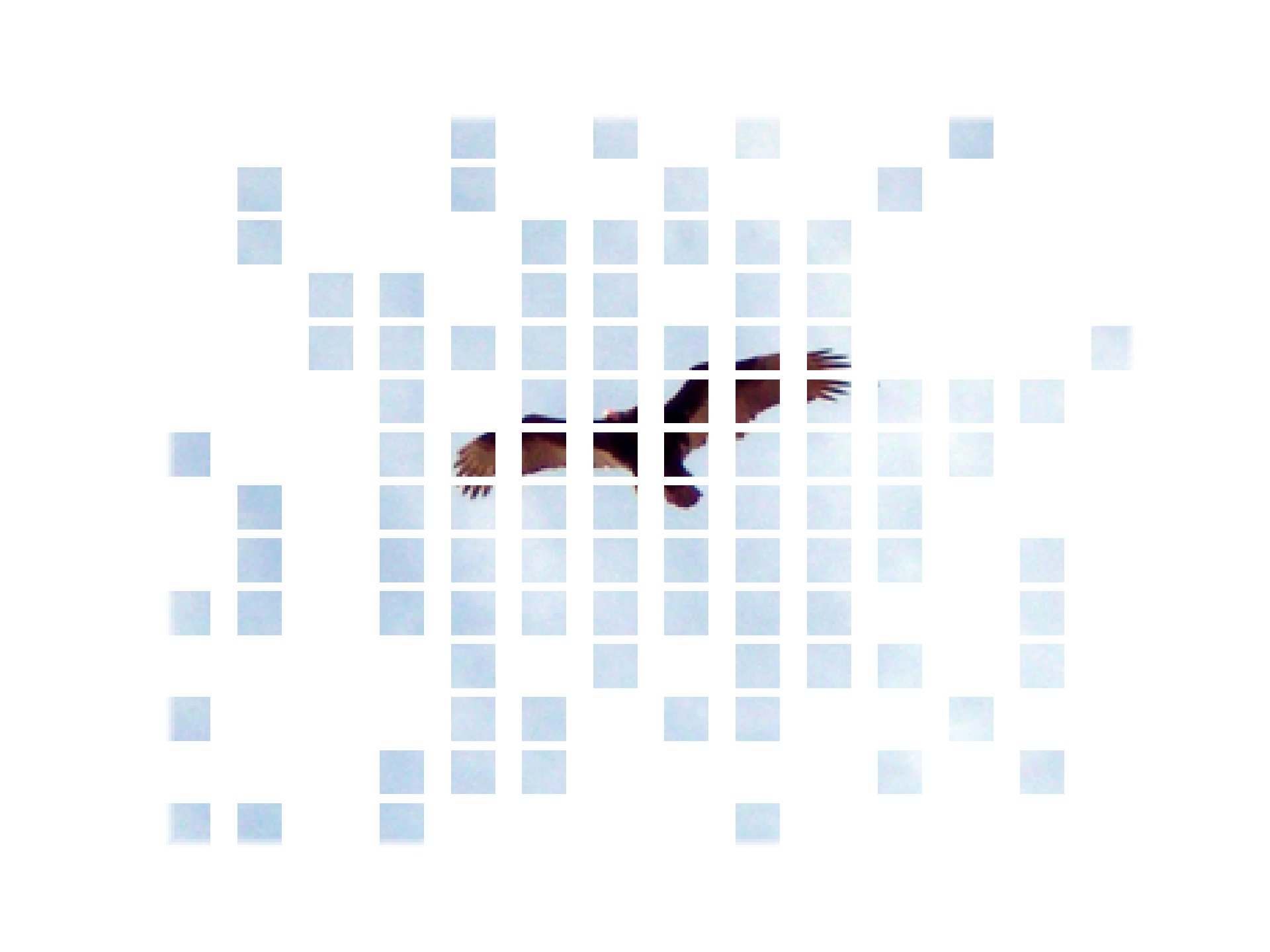}  
    \includegraphics[width=0.15\textwidth,height=0.15\textwidth]{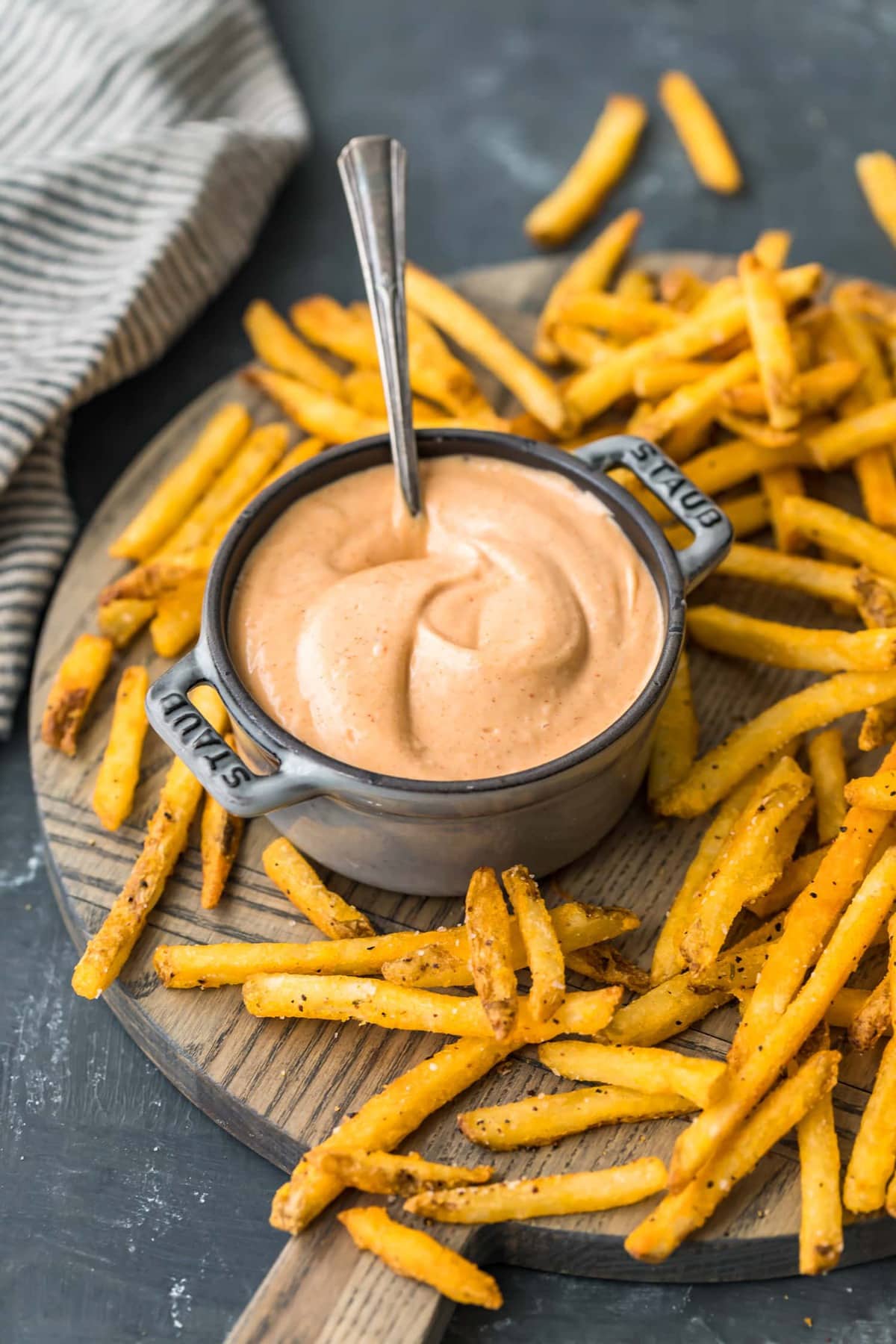} 
    \includegraphics[width=0.15\textwidth,height=0.15\textwidth]{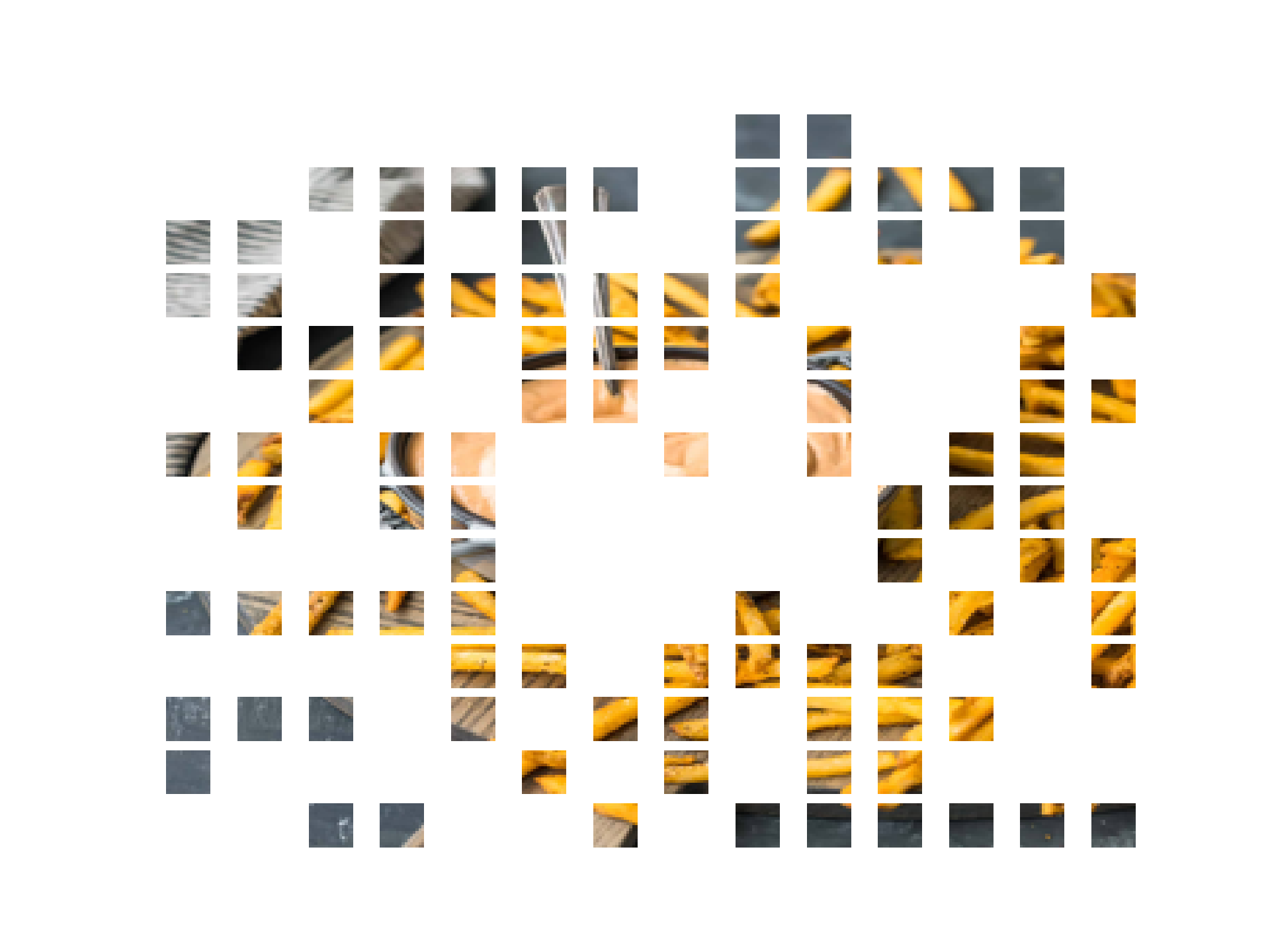} 
    \includegraphics[width=0.15\textwidth,height=0.15\textwidth]{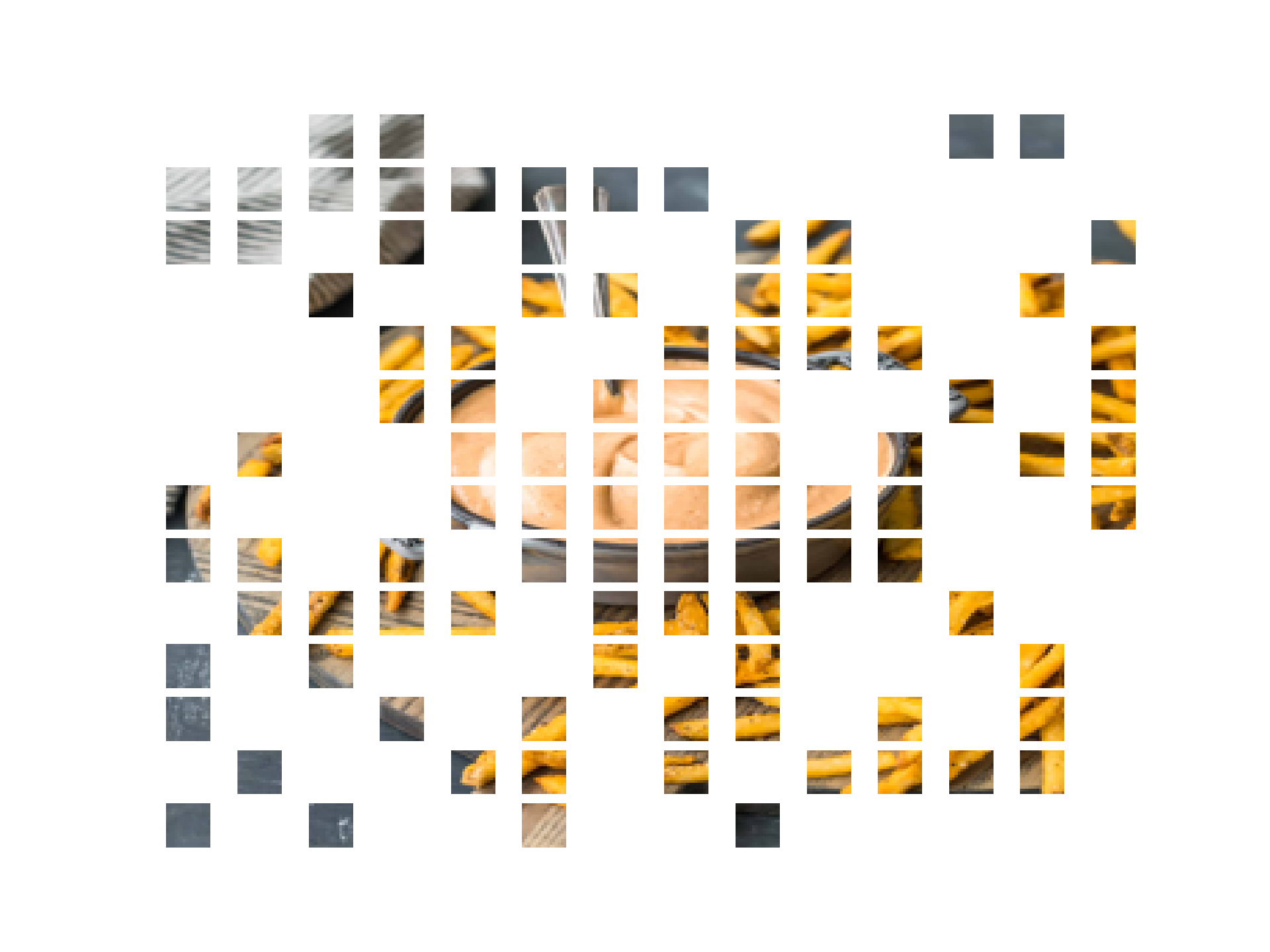}\\
     \includegraphics[width=0.15\textwidth,height=0.15\textwidth]{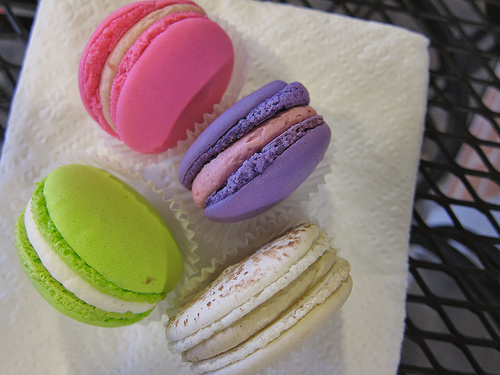} 
    & \includegraphics[width=0.15\textwidth,height=0.15\textwidth]{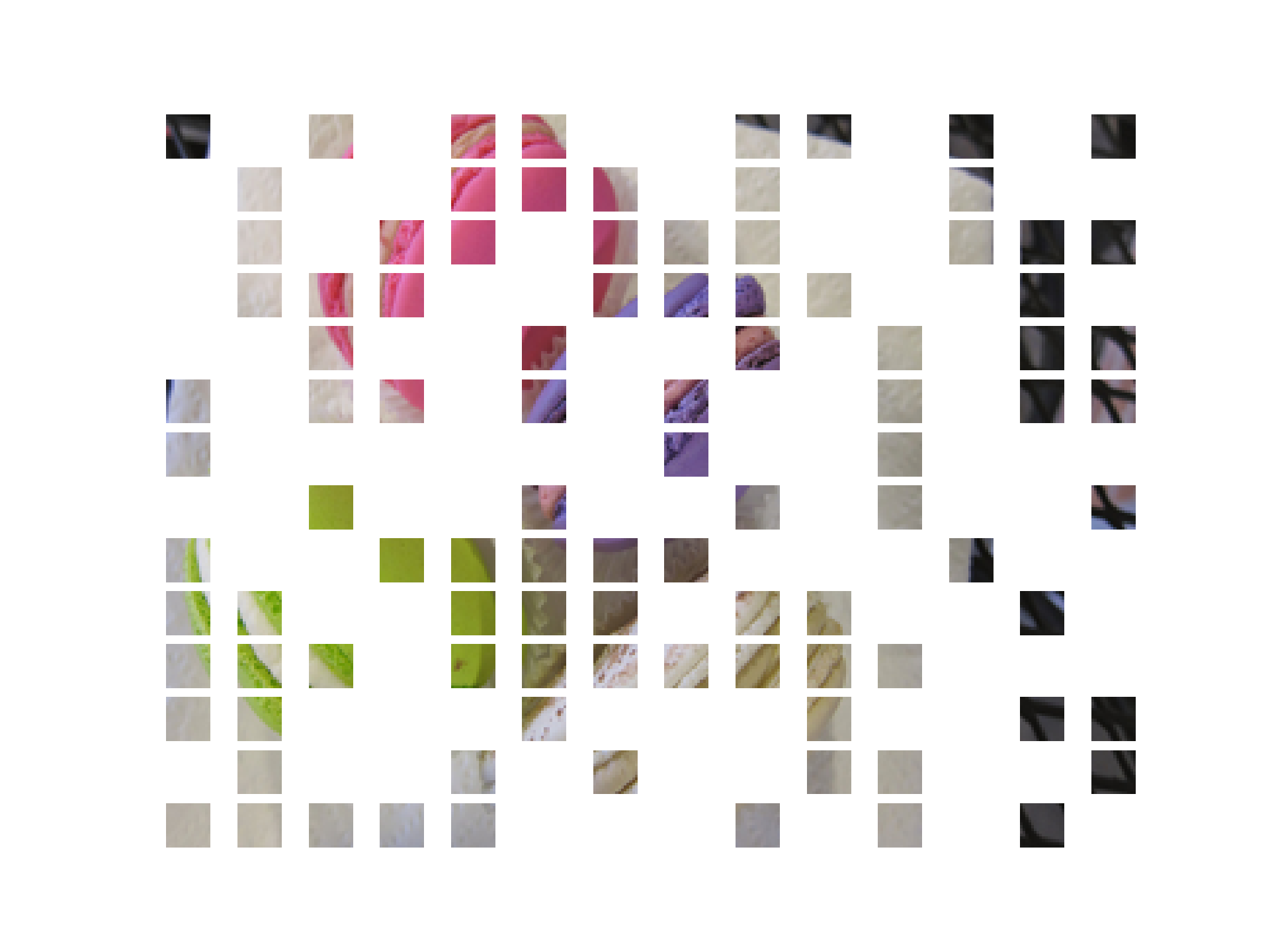}
    & \includegraphics[width=0.15\textwidth,height=0.15\textwidth]{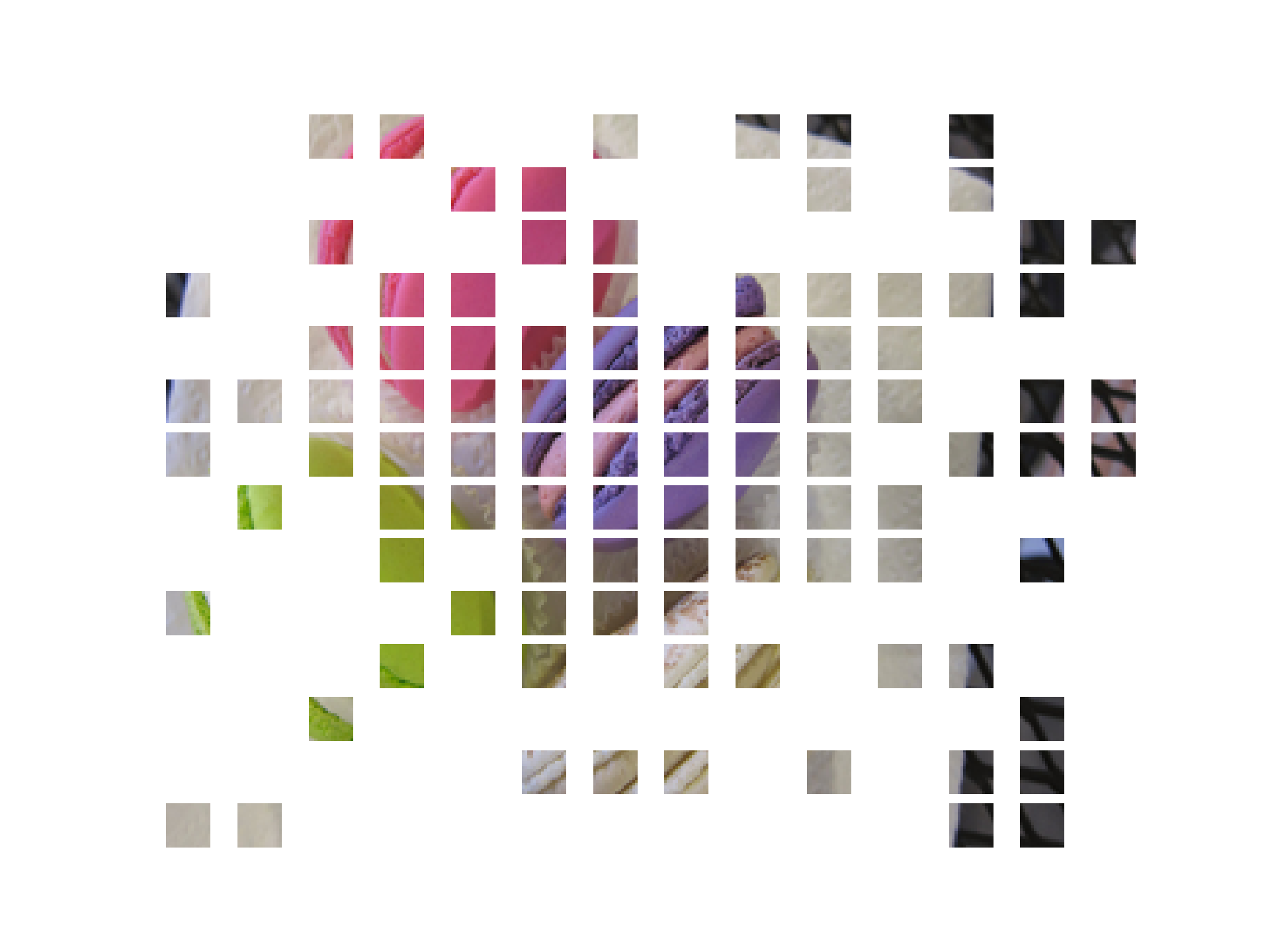}  
    \includegraphics[width=0.15\textwidth,height=0.15\textwidth]{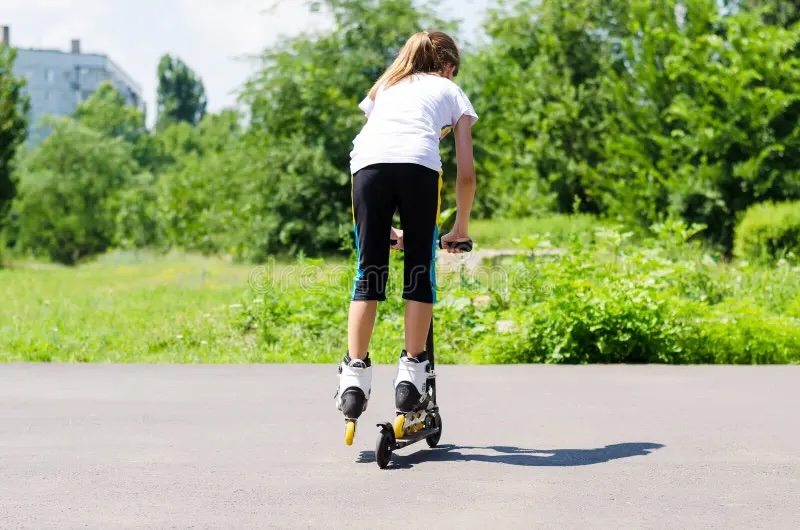} 
    \includegraphics[width=0.15\textwidth,height=0.15\textwidth]{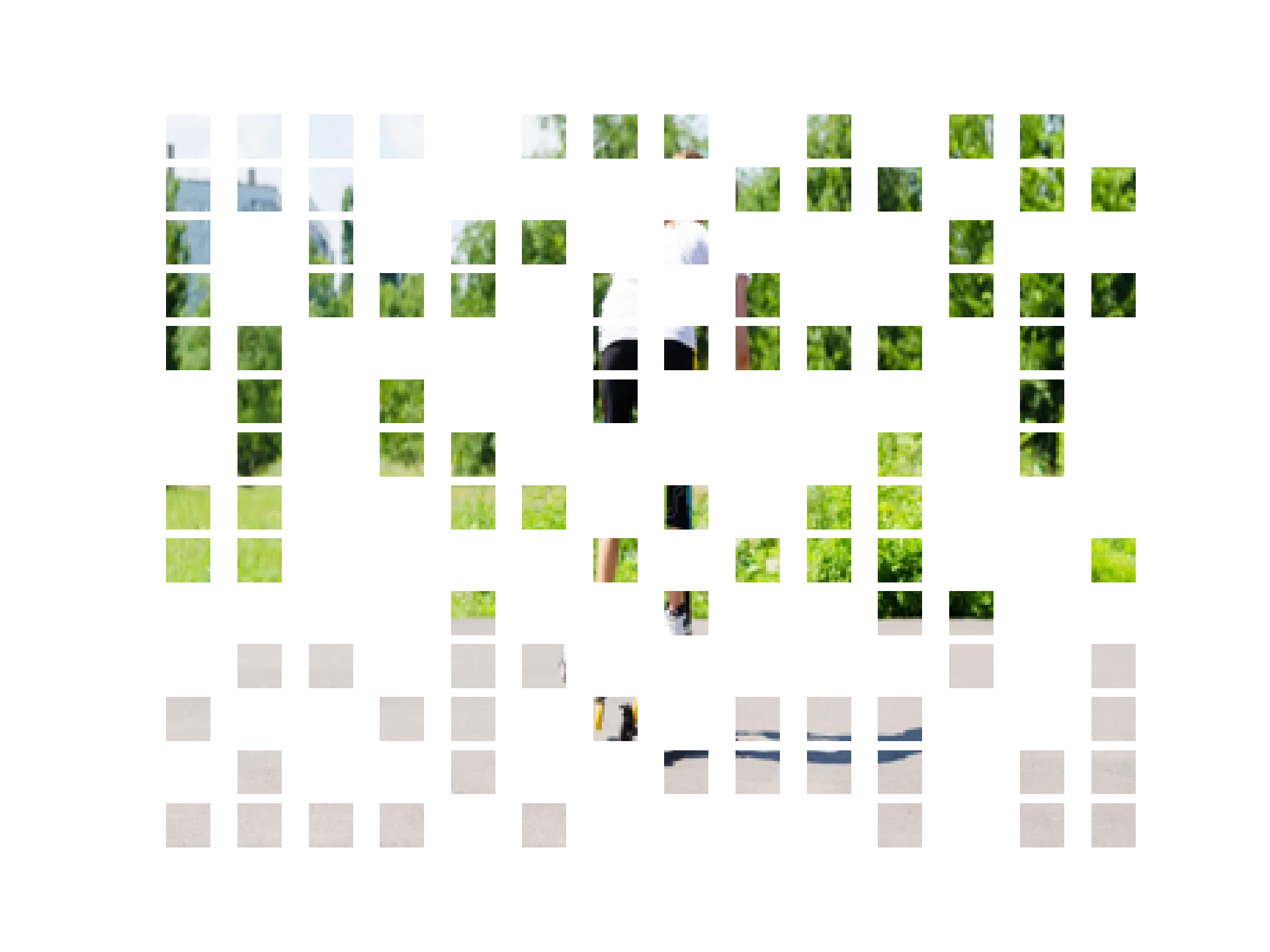} 
    \includegraphics[width=0.15\textwidth,height=0.15\textwidth]{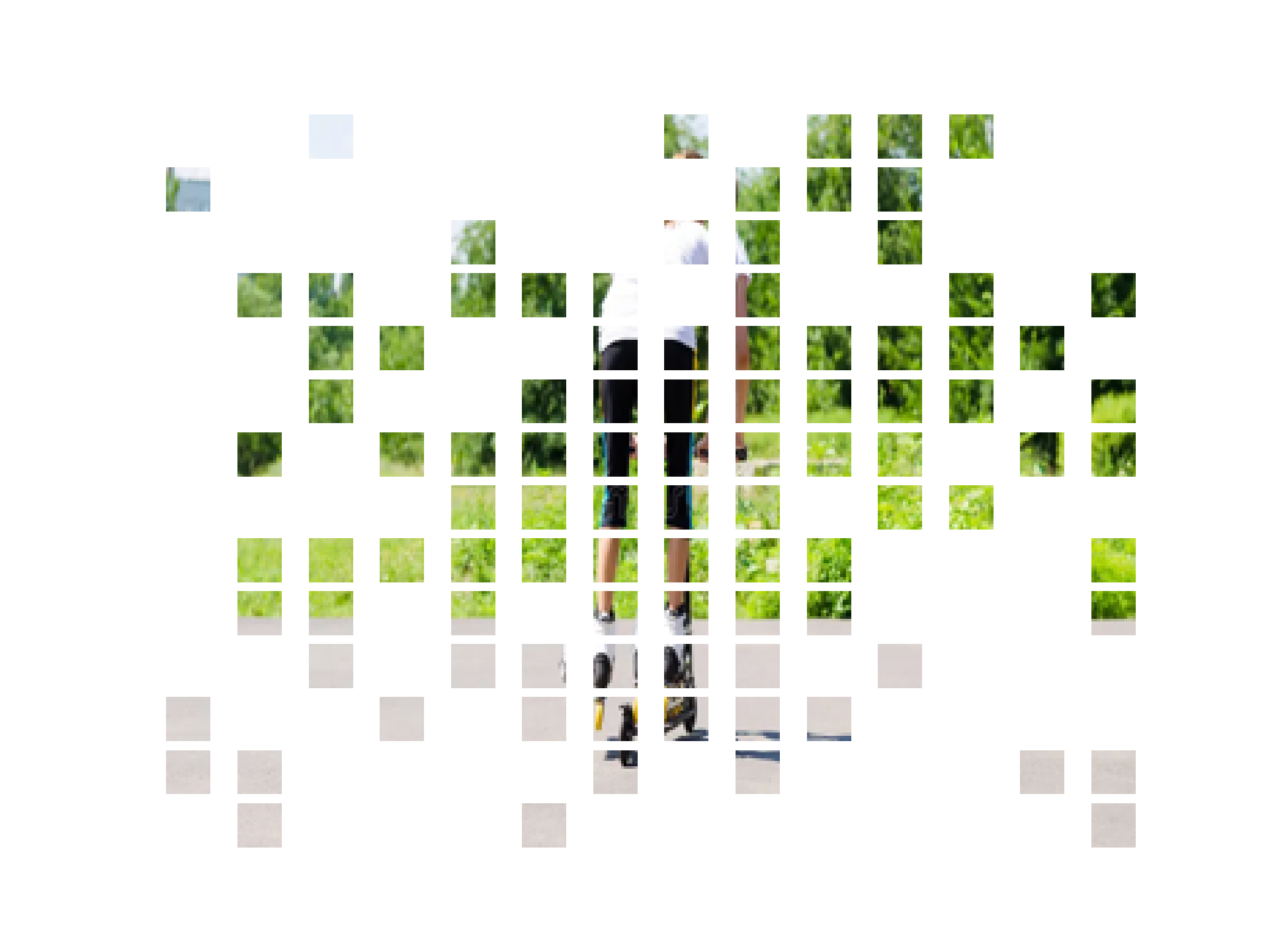} \\
    \end{tabular}
    \caption{
    Images from ImageNet-1K~\cite{deng2009imagenet} (left side) and CC12M~\cite{changpinyo2021cc12m} (right side).
    A contrast can be seen between random masking (left mask) and centered masking (right mask).
    Centered masking captures more of the main subject of the image.
    }
    \label{fig:exam}
\end{figure}

In this paper, we further improve on the performance of FLIP, while retaining the same savings in computation.
Specifically, we conjecture that patches closer to the center of an image are more important when training a Vision-Language Model than patches nearer the periphery.
Based on this conjecture, we propose an approach that replaces random masking of FLIP with centered masking.
Our approach, called Gaussian masking for Language-Image Pre-Training (GLIP), uses a Gaussian distribution centered at the middle of the image in order to select patches during training.
Figure~\ref{fig:exam} contrasts the random masking of FLIP with the centered masking of GLIP. 
It can be seen that the main subject of the image is better preserved in the retained patches when masking prioritizes patches from the center and disregards more patches at the periphery.  
The inspiration for the centered masking of GLIP comes from the behavior of human photographers, who often place the main subject of the photo in the center.
However, as will be seen from our experimental results, GLIP also delivers improvements for image data that does not have a strong photographic center focus, demonstrating GLIP's promise for generality. 
Note that we use masking in the same way that FLIP does, i.e., not for the purpose of reconstruction, which was shown by~\cite{li2023flip} to actually hurt performance.


GLIP can be considered to be related to other approaches that do not use random masking.
A key example is A-CLIP~\cite{yang2023aclip},
which employs an online Exponential Moving Average (EMA) to generate 
masks derived from the attention weights associated with the [\textit{CLS}] token of the visual encoder.
The attentive masks attempt to include more image patches that are highly correlated with the semantics of the text description.
GLIP aims to leverage the same effect, but instead of calculating attention, assumes that the most important part of the image is the center.
In this paper, we do not compare GLIP with A-CLIP because even the efficient version of A-CLIP requires more computational resources than GLIP. 

This paper makes the following contributions:
\begin{itemize}
    \item We introduce a centered masking approach for language-image pre-training called Gaussian Masking for Language-Image Pre-Training (GLIP), which enjoys the same reduction in computational resources as FLIP.
    \item We provide extensive experiments that show the superiority of GLIP over FLIP. GLIP shows general performance improvements and also allows higher masking ratios. 
    \item We show that GLIP delivers an easy-to-obtain improvement across different datasets without requiring effort to tune the variance of the Gaussian.  
\end{itemize}

Our code is available online\footnote{https://github.com/Anastasiais-ml/GLIP}. 
In our experiments comparing FLIP and GLIP we train on a smaller dataset than was used in the original FLIP paper~\cite{li2023flip}.
For this reason, the results of our paper can be easily reproduced using the resources typically available in either an academic or industry setting.

\section{Related work}
\subsection{Vision-Language Models}
Refining visual representations of images and developing foundation models to enhance subsequent tasks is a key objective in the field of computer vision.
CLIP~\cite{CLIP2021radford}, and closely related ALIGN~\cite{scaling2021Jia}, learn visual representations using natural language supervision.
The approach involves collecting billions of image-text pairs from the internet and pre-training the model on the data using contrastive learning techniques. 
This extensive pre-training on large-scale datasets substantially improves the transferability of the model to a variety of downstream tasks and datasets.

However, pre-training models on billion-scale datasets require thousands of GPU days, which is prohibitively expensive.
The Fast Language-Image Pre-training (FLIP) approach involves randomly masking a large proportion of the patches in each training image to improve the efficiency of pre-training.
Remarkably, as mentioned above, FLIP not only outperforms models trained without masking but also reduces computational requirements by 2--4 times by masking 50\% to 75\% of the image patches.

The Attentive Mask CLIP (A-CLIP) refines the image patch masking strategy by utilizing correlation scores between image and text~\cite{yang2023aclip}. 
The correlation scores are computed using the attention weights to the [CLS] token of the visual encoder. 
By reducing the image resolution of the Exponential Moving Average (EMA) encoder by half and masking 50\% of the image patches of the image encoder, A-CLIP reduces 14\% of pre-training time while also improving performance.
Nevertheless, selecting external patches for processing still requires over 30\% more training time compared to masking patches directly.

Masked autoencoders (MAE) develop an encoder-decoder architecture to visual representations through self-supervision~\cite{he2022mae}.
In this approach, the MAE randomly masks most of the patches in the input image during the encoding stage and then reconstructs these masked patches using the decoder.
The pre-trained encoders developed through this method can be effectively used in downstream tasks, such as image classification.
Although MAE appears to have originally inspired FLIP, the original FLIP paper~\cite{li2023flip} did not find reconstruction helpful, and GLIP also does not use reconstruction.

\subsection{Relative Importance of Image Regions}
There are several strands of literature motivating our move from random masking to centered masking because the build on the idea that not all parts of an image are equally important.
The first that comes to mind might be the concept of model attention in deep learning, which is a mechanism that allows models to focus on specific parts of the input data that are more relevant for a given task~\cite{bahdanau2014neural,Jan2015NIPS,lu2016hierarchical,mnih2014recurrent,NIU202148}.
GLIP exploits the same notion that underlies model attention but in a simple manner rooted in the assumption that the center of the image is most important.

The importance of the center of the image is exploited for data augmentation during training.
Center cropping and random cropping are a commonly used data augmentation technique for image preprocessing for deep learning models, especially in tasks related to computer vision~\cite{NEURIPS2020Cubuk,devries2017dataset,krizhevsky2012alexnet,shorten2019survey} and Vision-Language Model~\cite{li2023flip,CLIP2021radford}.
Center cropping assumes that the most relevant information is located in the center of the image and helps to reduce overfitting by providing the model with slightly different views of the same image during training~\cite{NEURIPS2020Cubuk,devries2017dataset,krizhevsky2012alexnet}.
This assumption applies to many datasets, especially those collected from the Internet, which contain images framed by photographers~\cite{changpinyo2021cc12m,schuhmann2022laionb,Schuhmann2021laion400m,sharma2018cc3m}.
MaskCLIP~\cite{Dong2023MaskCLIP} and Masked Siamese Networks (MSN)~\cite{assran2022msn} augment images through the masking of image patches in Vision Transformer (ViT)~\cite{dosovitskiy2020ViT}, subsequently aligning the representations of both masked and unmasked images. 
This process contributes to the improvement of the semantic performance of images by ensuring the semantic content of both masked and unmasked images is coherent.
In our study, we do not draw comparisons with these methods.
In our study, we do not compare our method with MaskCLIP. 
This is because MaskCLIP focuses on maintaining semantic consistency between masked and unmasked images to enhance image representation via self-supervised learning, rather than reducing pre-training time.

\subsection{Photographer behavior} 
When photographers take pictures they have a strong tendency to place the subject of the image near the center of the image~\cite{arnheim1954art,roy1979art,tatler2007central}.
The placement of the subject or object within a photograph greatly influences its composition and aesthetic appeal~\cite{arnheim1954art,roy1979art}. 
The Rule of Thirds and Center Composition are two widely accepted techniques for enhancing the visual appeal of an image~\cite{arnheim1954art}.
The Rule of Thirds involves dividing the frame into nine equal sections with two horizontal and two vertical lines and positioning the subject at the intersections or along these lines to create balanced, visually pleasing images~\cite{arnheim1954art}.
Centre composition focuses on placing the subject at the center of the picture, highlighting symmetry and prominence, a technique commonly used in portraiture and symmetrical scenes~\cite{roy1979art}.
A commonality across these techniques is the avoidance of placing objects at the edges of the image.
We refer to the tendency of a dataset to contain images with the main subject material as ``center focus''. 
In addition to being created by a photograph ``center focus'' can also arise when someone crops an image after it has been taken so that the main subject appears close to the center.



\begin{figure}[!t]
    \centering
    \includegraphics[width=0.6\linewidth]{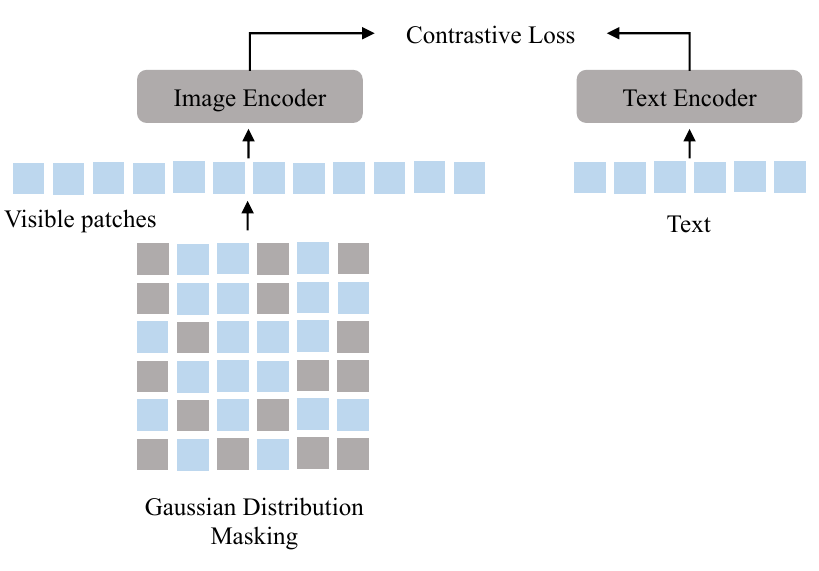}
    \caption{\textbf{Our GLIP architecture.} Following CLIP~\cite{CLIP2021radford} and FLIP~\cite{li2023flip}, we use contrastive loss to pre-train our model. 
    Different from FLIP, we mask image patches with a Gaussian distribution instead of random masking.}
    \label{fig:framework}
\end{figure}


\section{Method}

The key idea of FLIP is to enhance computational efficiency and increase the batch size of contrastive objectives by applying masks to the input image in CLIP. 
GLIP differs by masking image patches according to the idea that the center of an image is more important.
This technique substantially improves performance across various tasks almost without additional computation, outperforming FLIP.
Our image masking strategy is straightforward: we utilize probabilities from a Gaussian distribution to mask the image patches, as illustrated in Figure~\ref{fig:framework}. Considering the three-dimensional nature of the image, we employ a Bivariate Gaussian Distribution for this purpose:

\begin{equation}
    \resizebox{.9\hsize}{!}{$f(x, y) = \frac{1}{2\pi\sigma_x\sigma_y\sqrt{1-\rho^2}} \exp\left(-\frac{1}{2(1-\rho^2)}\left[\frac{(x-\mu_x)^2}{\sigma_x^2} + \frac{(y-\mu_y)^2}{\sigma_y^2} - \frac{2\rho(x-\mu_x)(y-\mu_y)}{\sigma_x\sigma_y}\right]\right)$}
\end{equation}

\begin{itemize}
    \item $f(x,y)$ is the probability density function of the bivariate Gaussian distribution.
    \item $\mu_x$ and $\mu_y$ are the means of the two variables, $x$ and $y$, respectively.
    \item $\sigma_x$ and $\sigma_y$ are the standard deviations of the two variables, $x$ and $y$, respectively.
    \item $\rho$ is the correlation coefficient between $x$ and $y$.
\end{itemize}

when set $\mu_x$ and $\mu_y$ to $0$ and set $\rho$ to $0$:
\begin{equation}
    f(x, y) = \frac{1}{2\pi\sigma_x\sigma_y} \exp\left(-\frac{1}{2}\left[\frac{x^2}{\sigma_x^2} + \frac{y^2}{\sigma_y^2}\right]\right)
    \label{equ:bnd}
\end{equation}

\begin{figure*}[!t]
\centering
    \begin{tabular}{cccc}
    \includegraphics[width=0.24\linewidth]{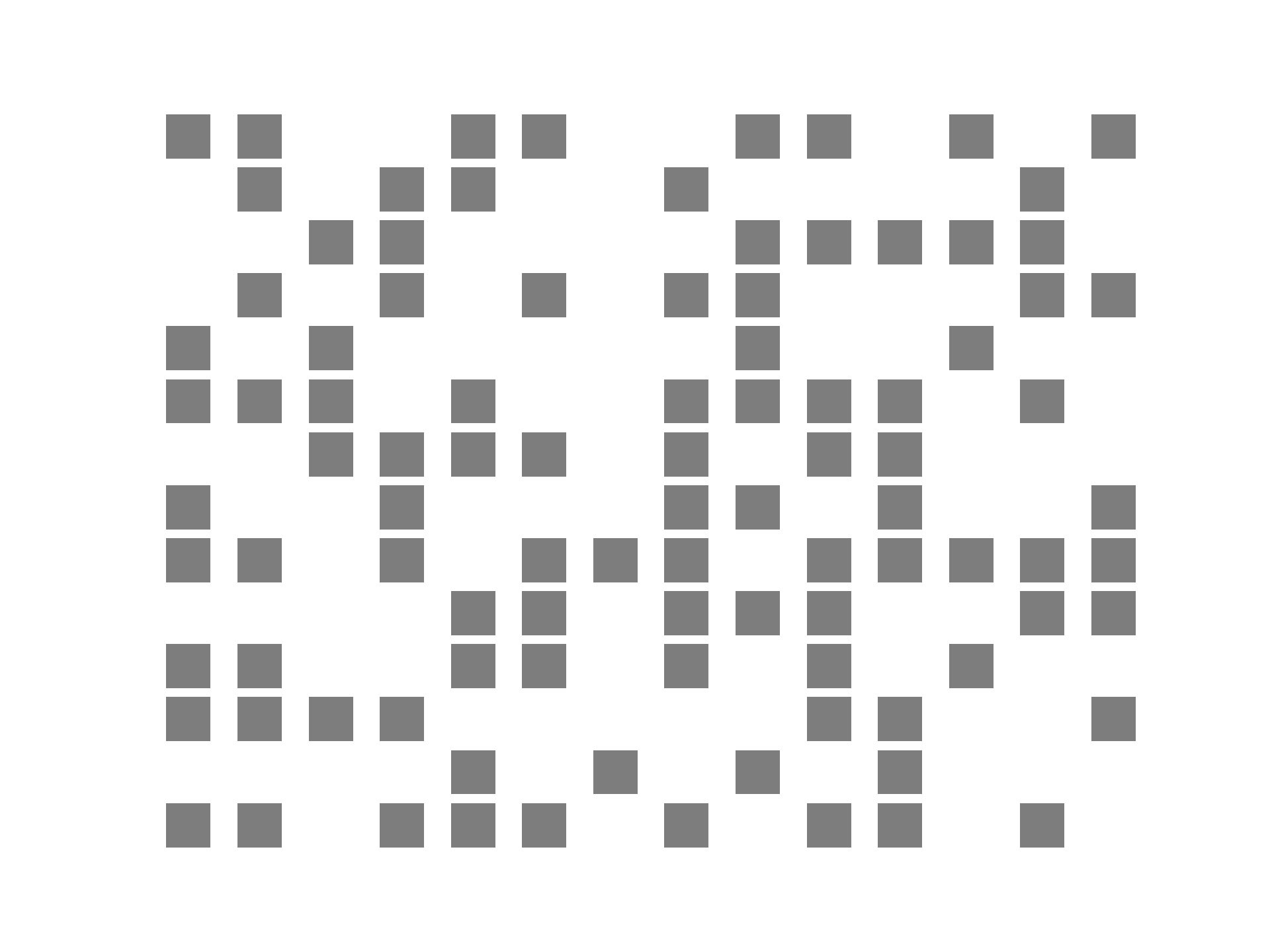} & \includegraphics[width=0.24\linewidth]{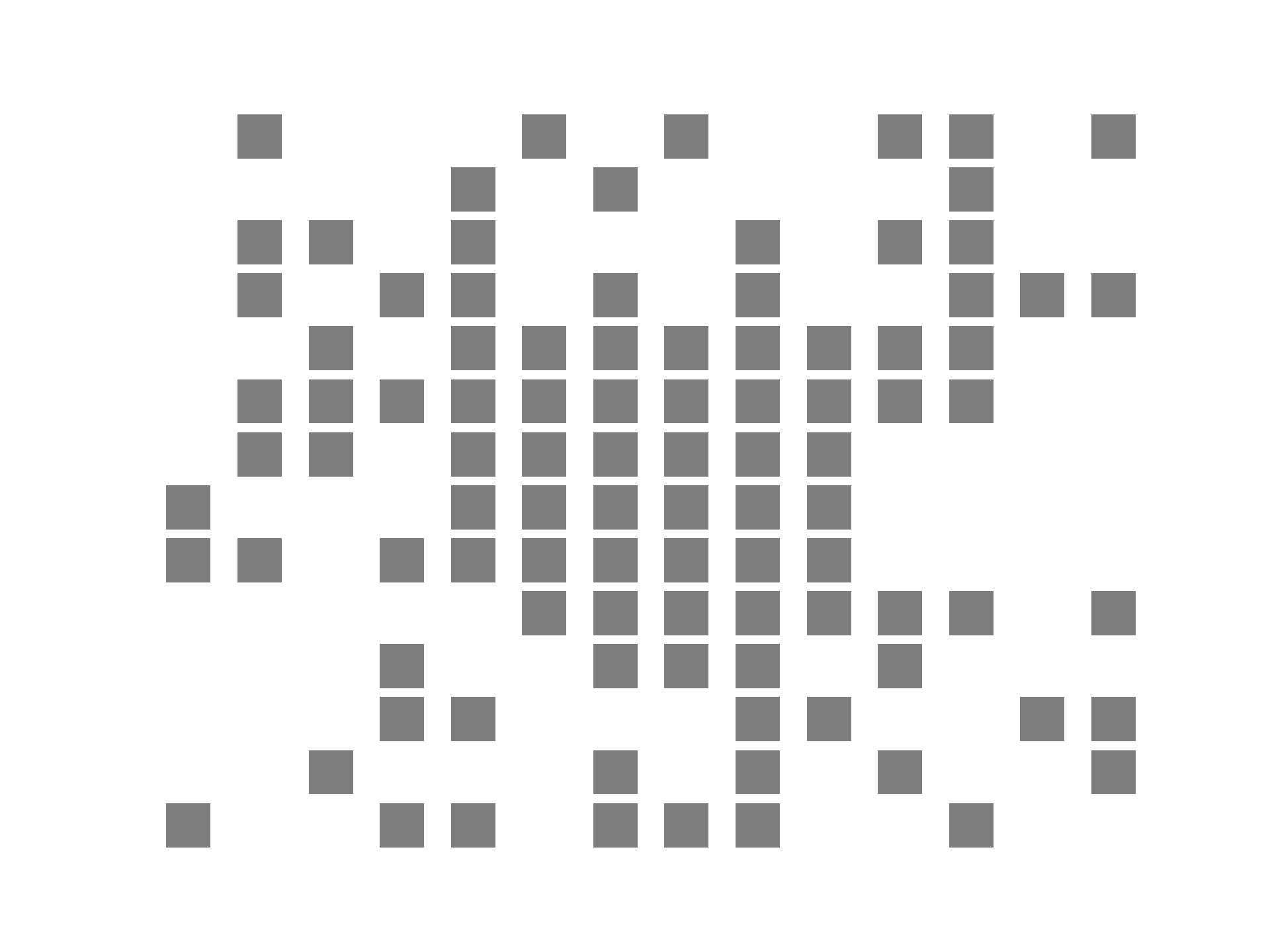} & \includegraphics[width=0.24\linewidth]{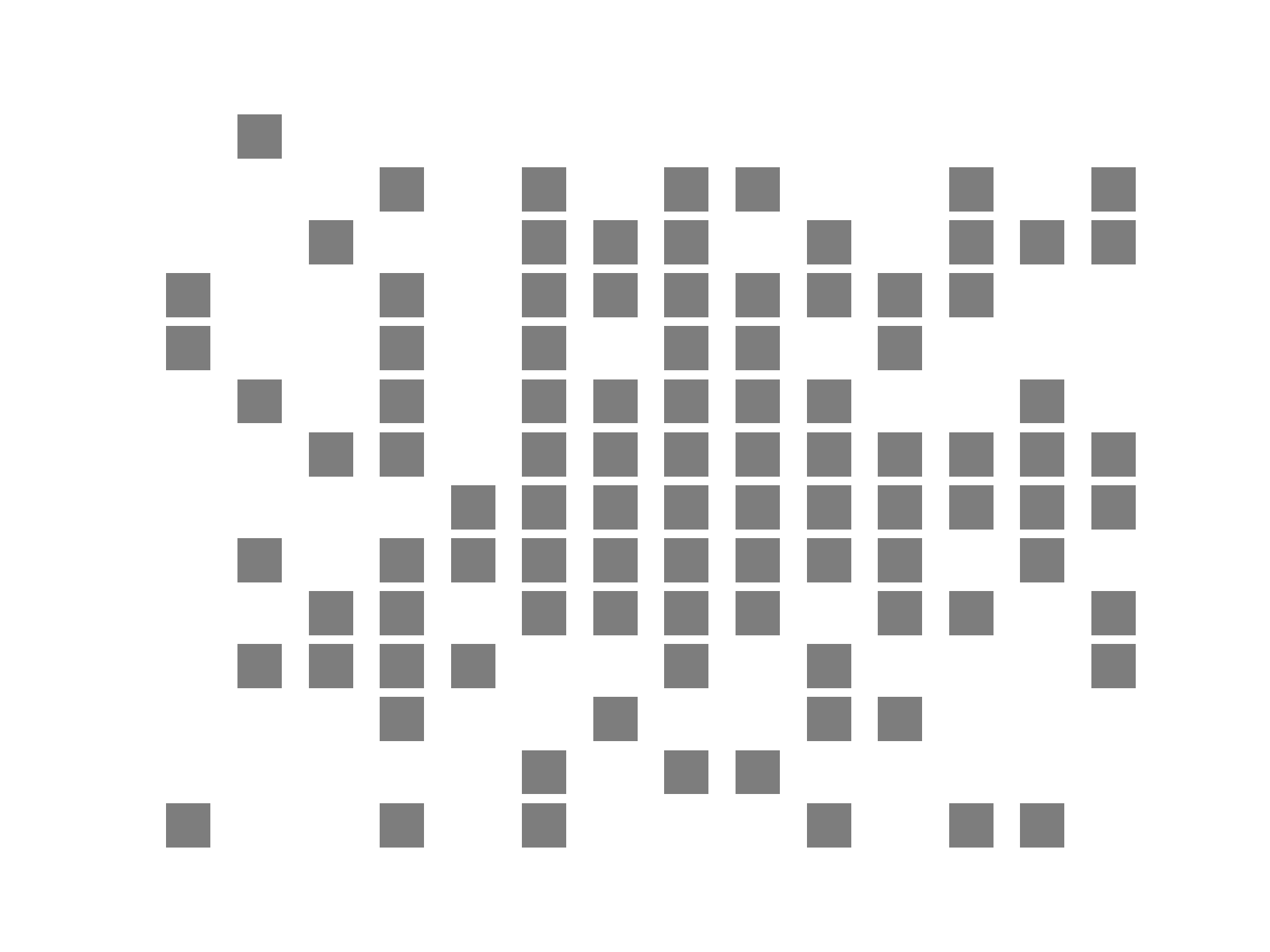} & \includegraphics[width=0.24\linewidth]{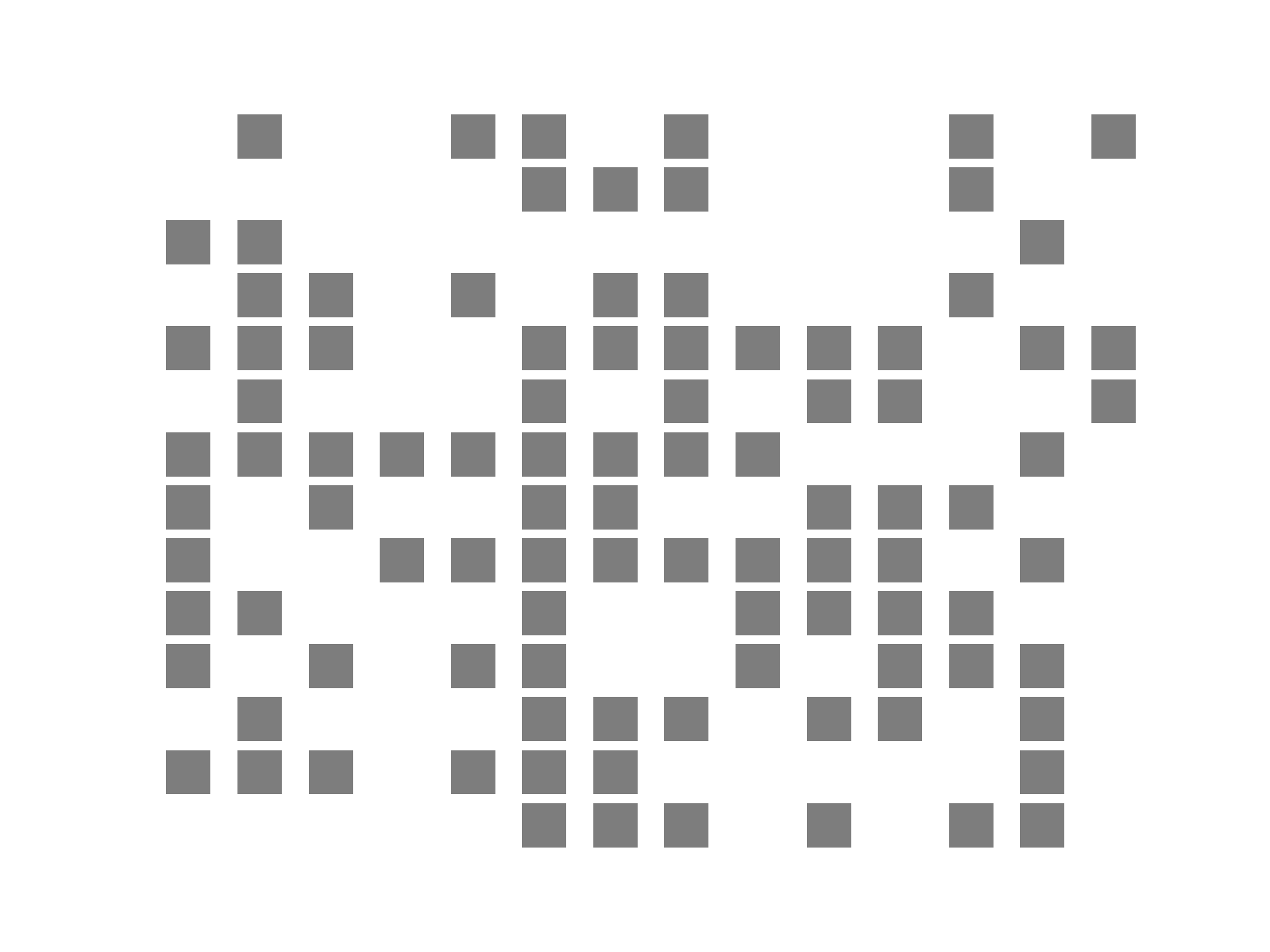} \\
    (a) Random Masking & (b) $\sigma=0.1$ & (c)  $\sigma=0.2$ & (d) $\sigma=0.8$ \\
    \end{tabular}
    \caption{\textbf{Comparison of Random and Gaussian Masking Strategies.} 
    Image (a) demonstrates a random masking strategy with uniform masking probability. 
    Images (b), (c), and (d) illustrate Gaussian masking with increasing standard deviations ($\sigma$),
    showcasing the effect of masking that is focused in the center and gradually spreads to the edges.
    }
\label{fig:random_gaussian_mask}
\end{figure*}

Finally, we use Equation~\ref{equ:bnd} to calculate the probability that an image patch will be masked.
In this formula, $x$ and $y$ are within the range of $[-1, 1]$, and the step size corresponds to the grid size of the image encoder.
We set the center of the image as the coordinate origin, which is the center of the bivariate Gaussian distribution.
The examples are shown in Figure~\ref{fig:exam} and ~\ref{fig:random_gaussian_mask}, we remain more patches in the center of the image.

Building on the CLIP~\cite{CLIP2021radford} method, we pre-trained the model using Noise-Contrastive Estimation (InfoNCE) loss with temperature parameter $\tau$~\cite{chen2020simple,CLIP2021radford} is shown in Equation~\ref{equ:infonce}.
This function works by bringing positive image-text pairs closer and negative image-text pairs farther apart~\cite{oord2018InfoNCE}.
This approach allows the encoder to learn and recognize similar semantics in the corresponding image-text pairs.

\begin{equation}
    \mathcal{L}_{\text{InfoNCE}} = -\log \left( \frac{\exp(\text{sim}(I_i, T_i)/\tau)}{\sum_{j=1}^{N}\exp(\text{sim}(I_i, T_j)/\tau)} \right)
    \label{equ:infonce}
\end{equation}
Here, $sim$ is the similarity score between image $I_i$ and text $T_i$.
Following FLIP~\cite{li2023flip}, we also continue pre-training the model without masking for a small number of steps to reduce the distribution gap between training and testing caused by masking.

\section{Experiments}

\begin{table}[t]
\centering
\caption{The details of pre-training and fine-tuning setup.}
    \begin{tabular}{l|l|l}
    Config & Pre-training Value & Fine-tuning Value \\
    \hline
    Optimizer & AdamW & AdamW \\
    Learning rate & 1e-3 & 5e-6 \\
    Weight decay & 0.1 & 0.05 \\
    Optimizer momentum & $\beta1, \beta2=0.9, 0.999$ & $\beta1, \beta2=0.9, 0.98$ \\
    Learning rate schedule & Cosine decay & Cosine decay \\
    Warmup steps & 10k & 10\% of total steps \\
    Epoch & 30 & 1 \\
    Numerical precision & Automatic mixed precision & Automatic mixed precision \\
    Augmentation & RandomResizedCrop & RandomResizedCrop\\
    \hline
    \end{tabular}
\label{tab:setup}
\end{table}

\subsection{Implementation Details}
Our implementation follows CLIP~\cite{CLIP2021radford}, OpenCLIP~\cite{Gabriel2021openclip} and FLIP~\cite{li2023flip}.

\textbf{Dataset.} 
We pre-trained our model on CC12M~\cite{sharma2018cc3m} dataset which includes 12 million image-text pairs.
To observe the performance of the model that was pre-trained on a different dataset, we further evaluate our method on CC3M~\cite{sharma2018cc3m}. 
These datasets were selected due to their diverse range of real-world concepts and scenes.
Models such as SLIP, CLIP, and BLIP~\cite{li2022blip,mu2021slip,yang2023aclip} also use these popular datasets to train their models.
Despite expired URLs, we successfully downloaded approximately 2.72 million and 9.30 million image-text pairs for CC3M and CC12M, respectively.
Notably, our model achieved similar baseline performance levels as reported by SLIP and A-CLIP for both CC3M and CC12M~\cite{mu2021slip,yang2023aclip}, demonstrating its effectiveness and robustness across diverse datasets.

\textbf{Architecture.}
For our image encoders, we employed the ViT-B/16~\cite{dosovitskiy2020ViT} with a patch size of 16, while utilizing a transformer-based model for the text encoder~\cite{Ashish2017Transformer}.
According to the results reported by FLIP, patch size is not an important contributor to performance.
The maximum context length of the text encoder used in our study is 32~\cite{li2023flip}.
Consistent with the configurations in CLIP~\cite{CLIP2021radford} and FLIP~\cite{li2023flip}, the input image size is 224 $\times$ 224.
In our principal experiments, we specified the values of \(\sigma_x\) and \(\sigma_y\) to be 0.20 in the formula~\ref{equ:bnd} as the default value, unless otherwise mentioned.
Additionally, we explored the impact of varying \(\sigma_x\) on the performance of GLIP.

\textbf{Training and fine-tuning.} 
Following the FLIP~\cite{li2023flip}, we pre-train the model for 30 epochs with patch masking and fine-tune the model for small steps without patch masking.
Pre-training and fine-tuning configurations are detailed in Table~\ref{tab:setup}, utilizing 8 RTX A5000 GPUs with batch sizes of 160, 320, 640, and 1024 for mask ratios of 0\%, 50\%, 75\%, and 90\%, respectively.

\subsection{Evaluation.}

First, we evaluate our approach using the ImageNet dataset for the zero-shot classification task, comparing its performance with CLIP and FLIP. 
Subsequently, we extended our evaluation to include various other datasets.
we follow the prompt engineering of CLIP~\cite{CLIP2021radford} and OpenCLIP~\cite{Gabriel2021openclip}.
We use the average of a set of 80 templates embedding for each class. 
Then, the cosine similarity between the caption and the image embeddings is calculated to obtain the top-1 class as the predicted class for each image.

Additionally, we evaluate our approach across various tasks to demonstrate its advantages. 
These tasks include image-text retrieval on the MS COCO~\cite{lin2014coco} and Flickr30K~\cite{young2014Flickr30k} datasets, as well as linear probing and fine-tuning on ImageNet-1K~\cite{deng2009imagenet}. 
This comprehensive evaluation highlights the versatility and effectiveness of GLIP in addressing the various challenges in the field.

\begin{table*}[!t]
  \centering
  \caption{\textbf{Zero-shot accuracy on ImageNet-1K classification.}
     We pre-trained the models for 30 epochs on the CC12M~\cite{changpinyo2021cc12m} dataset by different image patch mask ratios with ViT-B/16 as the image encoder. Then, we fine-tuned them by an additional epoch without image masking.
     }
  \begin{tabular}{c|c|cc|c}
      \hline
       \multirow{2}{*}{Method}   &  \multirow{2}{*}{Masking} & \multicolumn{2}{c|}{Inference} & \multirow{2}{*}{After tuning}\\
                 &         & Masking       & Unmasking    &  \\
      \hline
      CLIP     &   \xmark    & \xmark & 35.5 & \xmark  \\
      \hline
      FLIP     &   \multirow{2}{*}{50\%}    & 32.1 & 34.0 & 34.2  \\
      GLIP   &                          & 33.2 & \textbf{35.1} & \textbf{35.4} \\
      \hline
      FLIP     &   \multirow{2}{*}{75\%}    & 26.3 & 29.4 & 30.0 \\
      GLIP   &                          & 28.8 & 32.1 & 32.2 \\
      \hline
      FLIP     &   \multirow{2}{*}{90\%}    & 16.6 & 21.1 & 22.0 \\
      GLIP   &                          & 20.7 & 16.5 & 25.8 \\
      \hline
    \end{tabular}
    \label{tab:main_vit}
\end{table*}

\textbf{ImageNet zero-shot transfer.} 
We initially compare our model against the CLIP baseline on the ImageNet-1K dataset for zero-shot transfer tasks. 
As shown in Table~\ref{tab:main_vit}, our model, with a 50\% masking rate and inference unmasking, achieved a 35.1\% accuracy, slightly trailing CLIP by 0.4\%. 
However, after additional tuning without image unmasking, our model performance is very close to CLIP.
In comparison, FLIP's performance remains 1.3\% below CLIP's, even after unmasking tuning.
Subsequently, we evaluated GLIP against FLIP, also as shown in Table~\ref{tab:main_vit}. 
Here, GLIP surpassed FLIP by 1.1\% in accuracy, employing a 50\% masking rate and inference unmasking. 
After fine-tuning, our model maintained a 1.2\% lead over FLIP. 
Further, as we increase the masking ratio, GLIP's advantage over FLIP extends from 1.2\% (with a 50\% masking ratio and tuning) to 3.8\% (with a 90\% masking ratio and tuning), showcasing our approach's efficacy across different masking ratios. 

At a 90\% masking rate, GLIP exceeds FLIP's performance by 4.1\% with inference masking. 
However, performance drops substantially from 20.7\% to 16.5\% with inference unmasking, suggesting a reduction in the model's generalizability when masking is removed. 
Nevertheless, small steps of fine-tuning can substantially improve performance, from 16.5\% to 25.8\%, indicating that GLIP is better at learning the relationship between images and text compared to FLIP.
Impressively, GLIP also achieved a zero-shot classification accuracy of 25.8\% on ImageNet-1K, while requiring only 10\% of the image encoder computing resources normally required by an image encoder.
Large masking ratios allow us to use larger batch sizes to pre-train the model.
We anticipate that GLIP will effectively extend the advantages observed on smaller datasets to larger ones, such as LAION-400M~\cite{Schuhmann2021laion400m} and LAION-2B datasets~\cite{schuhmann2022laionb}.
This is because larger mask ratios allow more samples to be seen in the same training time, a factor that has been proven to enhance performance in the CLIP~\cite{CLIP2021radford} and OpenCLIP~\cite{Gabriel2021openclip}.


\begin{table}[!t]
\centering
\caption{\textbf{Linear probing and fine-tuning accuracy on ImageNet-1K classification.}
The performance of FLIP and GLIP that pre-trained on CC12M~\cite{changpinyo2021cc12m}. 
}
\begin{tabular}{ll|ccc}
\hline
    Dataset & Method & 0-shot & Linear & Finetuned \\ \hline
    \multirow{3}{*}{CC12M}   & CLIP     & 35.5 & 59.87 & 71.80 \\
                             & FLIP     & 34.2 & 59.13 & 71.75 \\ 
                             & GLIP     & 35.4 & 59.97 & 71.80 \\ \hline
\end{tabular}
\label{table:pb_ft}
\end{table}

\textbf{ImageNet linear probing.} 
Linear probing is to fine-tune a linear classifier on top of a pre-trained model while keeping the model's weights fixed.
The second column of Table~\ref{table:pb_ft} shows the performance outcomes of various methods pre-trained on different datasets. 
In this evaluation, GLIP surpasses the performance of FLIP, achieving an improvement of 0.84\%. Additionally, GLIP marginally outperforms CLIP, with a 0.1\% higher accuracy.

\textbf{ImageNet fine-tuning.}
For ImageNet fine-tuning, the process includes training linear classifiers on the ImageNet dataset without freezing the pre-trained model's weights. 
In this case, CLIP and GLIP perform equally well on this task, slightly better than FLIP, with a 0.05\% lead over FLIP.


\begin{table}[!t]
    \centering
    \caption{\textbf{Zero-shot accuracy on more classiﬁcation datasets}.
    FLIP and GLIP are pre-trained with 50\% ratio image patch masking and fine-tuned an additional epoch without image masking.
    }
    \begin{tabular}{l|l|c|cc|cc}
        \hline
        \multirow{2}{*}{\textbf{Category}} &  \multirow{2}{*}{\textbf{Dataset}} & \multirow{2}{*}{\textbf{CLIP}} & \multicolumn{2}{|c|}{Before tuning} & \multicolumn{2}{c}{After tuning} \\
        \cline{4-7}
                    & & & \textbf{FLIP} & \textbf{GLIP} & \textbf{FLIP} & \textbf{GLIP} \\ \hline
        \multirow{13}{*}{Object} 
        & Food101     & 40.46 & 39.02 & \textbf{40.04} & 39.68 & \textbf{40.42} \\
        & CIFAR10     & 66.22 & \textbf{59.41} & 51.63 & \textbf{56.19} & 49.24 \\
        & CIFAR100    & 28.83 & 26.10 & \textbf{26.13} & 24.59 & \textbf{24.86} \\
        & CUB         & 7.28  & \textbf{9.42}  & 9.23  & 9.79  & \textbf{9.98}  \\
        & Cars        & 13.70 & 10.24 & \textbf{13.02} & 10.24 & \textbf{12.78} \\
        & Aircraft    & 2.38  & 2.59  & \textbf{3.11}  & 2.47  & \textbf{3.05}  \\
        & DTD         & 18.56 & 16.76 & \textbf{18.14} & 17.02 & \textbf{18.30} \\
        & Oxford Pets & 51.67 & 50.06 & \textbf{54.30} & 50.33 & \textbf{54.95} \\
        & Caltech101  & 71.06 & 66.82 & \textbf{69.72} & 65.99 & \textbf{70.16} \\
        & Flowers102  & 1.37  & 2.48  & \textbf{3.47}  & 1.60  & \textbf{3.51}  \\
        & MNIST       & 14.46 & \textbf{9.80}  & 9.36  & \textbf{9.80}  & 9.74  \\
        & STL10       & 91.00 & 87.25 & \textbf{88.75} & \textbf{87.70} & 87.64 \\
        & GTSRB       & 10.10 & \textbf{13.24} & 7.95  & \textbf{12.75} & 9.07  \\
        & ImageNet-1K & 35.51 & 33.95 & \textbf{35.12} & 34.22 & \textbf{35.35} \\
        \hline
        \multirow{10}{*}{Others} 
        & SUN397      & 49.77 & 47.69 & \textbf{48.10} & 47.35 & \textbf{47.94} \\
        & EuroSAT     & 23.92 & 12.48 & \textbf{20.20} & 12.18 & \textbf{19.38} \\
        & RESISC45    & 34.84 & \textbf{33.65} & 32.59 & \textbf{33.84} & 33.10 \\
        & Country211  & 4.36  & \textbf{4.18}  & 3.97  & \textbf{4.29}  & 4.03  \\
        & PCam        & 50.62 & 50.21 & \textbf{57.22} & 52.41 & \textbf{55.94} \\
        & KITTI       & 25.67 & \textbf{37.30} & 34.22 & \textbf{37.90} & 30.95 \\
        & UCF101      & 40.10 & 37.67 & \textbf{38.09} & 37.48 & \textbf{38.33} \\
        & Kinetics700 & 24.07 & 23.15 & \textbf{23.70} & 23.15 & \textbf{23.51} \\
        & CLEVR       & 18.48 & 20.58 & \textbf{21.39} & 18.77 & \textbf{21.48} \\
        & HatefulMemes& 54.36 & \textbf{51.64} & 50.59 & \textbf{51.01} & 50.76 \\
        & SST2        & 50.14 & \textbf{50.41} & 49.48 & \textbf{50.58} & 49.37 \\
        \hline
    \end{tabular}
    \label{table:sorted_dataset_comparison}
\end{table}

\textbf{Zero-shot classification on more datasets.}
We evaluate our method across a variety of datasets to demonstrate its superiority. 
These are the same datasets that were used in the original CLIP~\cite{CLIP2021radford} and FLIP~\cite{li2023flip} papers.
In order to investigate whether GLIP has a dependence on the characteristics of data, we divide the datasets into two groups: ``Object'' and ``Other''.
Images of ``Object'' datasets depict an object and are framed so that the object is in the middle of the image nearly without exception.
Images of the ``Other'' datasets include scenes, activities, and datasets that otherwise cannot be considered ``Object'' datasets.
Note that the ``Other'' datasets might also have a center focus, meaning that the main subject matter occurs towards the center of the image, although this subject material might not be an object (e.g., it might be a set of objects in a scene).
According to our scan of the content of these datasets, the two datasets that are neither photographer framed nor cropped to center the main subject content and clearly depart from the assumption of center focus are EuroSat (satellite images)~\cite{helber2019eurosat} and PCAM (medical images)~\cite{ehteshami2017pcam,Veeling2018PCam}.

In Table~\ref{table:sorted_dataset_comparison},  
we see that GLIP outperforms FLIP in 
in the majority of ``Object'' datasets both before and after fine-tuning.
Fine-tuning in general provides a small but noticeable advantage in some but not all cases.
Recall that the number of patches used by GLIP is the same as FLIP, and the only difference is that GLIP prefers to retain patches closer to the center during masking.

Moving to the ``Other'' datasets, we might expect that GLIP is less helpful, since these images are less likely to have their main subject materially located at the center of the image.
We see that GLIP still delivers an improvement over FLIP in relatively many cases.
Surprisingly, this improvement appears to be greatest for EuroSAT and PCam, the two datasets that are clearly not center focused.
These results are interesting because they demonstrate that the potential of GLIP is not restricted to image data with specific characteristics.
We return to comment on the cases of EuroSAT and PCam in the outlook. 

In sum, before fine-tuning, 
The overall average absolute performance of GLIP on the 26 downstream datasets is improved by 0.54\% compared to FLIP.
Following fine-tuning, GLIP continues to hold its advantage across these datasets, underscoring its robustness and efficiency in handling a variety of datasets. 

It is worth noting that both FLIP and GLIP do not perform as well as CLIP on low-resolution datasets, such as MNIST, CIFAR-10, CIFAR-100, STL10~\cite{adam11stl10,krizhevsky2009cifar,lecun1998mnist}.
The original FLIP paper~\cite{li2023flip} does not observe this issue.
For our smaller training set, it possibly could have been addressed by data augmentation, a point we leave for future work.
It is also important to note that in contrast to the original FLIP paper~\cite{li2023flip}, our FLIP model (50\% sampling ratio) does not consistently outperform CLIP. 
This suggests that very large batch sizes are critical for optimal performance of the sampling approaches and deserve further investigation.

\textbf{Zero-shot robustness evaluation}
In Table~\ref{tab:imagenet_all}, we evaluate robustness, following the methodologies of Table 16 in~\cite{CLIP2021radford}. 
GLIP surpasses FLIP on 4 out of 6 datasets by a margin of 1.15\% and achieves comparable performance to CLIP which is pre-trained on the entire images. 
This comparison highlights GLIP's effectiveness in enhancing robustness across diverse datasets.

\begin{table}[!t]
\centering
\caption{\textbf{Zero-shot robustness evaluation}, comparison the zero-shot accuracy performance of CLIP, FLIP, and GLIP on various datasets. 
}
\begin{tabular}{l|c|cc|cc}
    \hline
     \multirow{2}{*}{Dataset} &  \multirow{2}{*}{CLIP} & \multicolumn{2}{c}{Before tuning} & \multicolumn{2}{|c}{After tuning}\\
                        &   & FLIP   & GLIP & FLIP  & GLIP\\ 
    \hline
    ImageNet-A      &  8.47 & \textbf{7.03}   & 6.95     & \textbf{7.47}	&   7.45 \\
    ImageNet-O      & 37.45 & \textbf{39.80}  & 38.75    & \textbf{39.85} &	38.95 \\
    ImageNet-R      & 46.01 & 39.59  & \textbf{43.06}    & 40.24 &	\textbf{43.47} \\
    ImageNet Sketch & 23.57 & 19.97  & \textbf{22.03}    & 20.24 &	\textbf{22.20} \\
    ImageNetV2      & 30.07 & 29.09  & \textbf{30.12}    & 29.24 &	\textbf{30.47} \\
    ObjectNet       & 21.85 & 18.38  & \textbf{19.32}    & 19.56 &	\textbf{20.98} \\
    \hline
    Average	        & 27.90	& 25.64	 & \textbf{26.71}	& 26.10	& \textbf{27.25} \\
    \hline
\end{tabular}
\label{tab:imagenet_all}
\end{table}


\begin{table*}[!t]
\centering
\caption{\textbf{Zero-shot Image-Text Retrieval,} we evaluated CLIP, FLIP, and GLIP image-text retrieval performance on COCO and Flickr30k datasets.
FLIP and CLIP are pre-trained with 50\% ratio image patch masking and fine-tuned an additional epoch.}
\resizebox{\textwidth}{!}{%
\begin{tabular}{l|lll|lll|lll|lll}
\hline
   \multirow{3}{*}{Model}  & \multicolumn{6}{c|}{Text Retrieval}    & \multicolumn{6}{c}{Image Retrieval}    \\
                        & \multicolumn{3}{c|}{Flickr30k} & \multicolumn{3}{c|}{COCO} & \multicolumn{3}{c|}{Flickr30k} & \multicolumn{3}{c}{COCO} \\
                        & R@1      & R@5     & R@10     & R@1    & R@5    & R@10   & R@1      & R@5     & R@10     & R@1    & R@5    & R@10   \\
                       \hline
                       CLIP        & 55.40 & 81.70 & 88.60 & 34.42 & 61.62 & 72.16 & 41.52 & 70.76 & 79.84 & 23.65 & 47.64 & 59.13 \\
                       FLIP        & \textbf{53.90} & 79.70 & \textbf{87.50} & 31.82 & 59.12 & 70.42 & 39.80 & 67.58 & \textbf{76.66} & 22.42 & 45.62 & 57.40 \\
                       \hline
                       GLIP    & 53.70 & \textbf{80.80} & 86.60 & \textbf{32.74} & \textbf{59.18} & \textbf{71.30}  & \textbf{41.92} & \textbf{67.66} & 76.42 & \textbf{22.53} & \textbf{46.61} & \textbf{58.52} \\
                       \hline    
\end{tabular}}
\label{tab:zero-shot retrieval}
\end{table*}

\textbf{Image-Text retrieval}
Table~\ref{tab:zero-shot retrieval} presents the performance of image-text retrieval on the COCO~\cite{lin2014coco} and Flickr30K~\cite{young2014Flickr30k} datasets. 
GLIP outperforms FLIP on both datasets. 
However, the performance of both GLIP and FLIP falls short of that achieved by CLIP, which was pre-trained on the CC12M dataset without any masking. 
This contrasts with previous reports suggesting that FLIP surpasses CLIP in zero-shot image-text retrieval tasks~\cite{li2023flip}.
With these models being pre-trained on a 400M dataset, our method may still hold an advantage when applied to larger datasets.
Notably, GLIP demonstrates a substantial advantage over FLIP in image-text retrieval tasks.

\textbf{Inverse Gaussian Masking}
We carry out an experiment in which we invert the Gaussian mask used by GLIP in order to confirm the importance of the center patches of the image.
We ``inverse GLIP'' a standard task, i.e., ImageNet-1K zero-shot classification.
As expected, this strategy performs poorly (as shown in row 2 Table~\ref{table:rec}), GLIP, which employs center masking, outperforms inverse Gaussian masking by 4.5\%.
Additionally, inverse Gaussian masking falls 3.3\% short of FLIP, which utilizes random masking. 
Because ImageNet-1K is photographer framed, we attribute
this difference in performance to the periphery of an image being more likely to contain background information rather than the main subject of the image.
The contrast between the performance of inverse-GLIP and FLIP suggests that the strength of FLIP actually derives from those random patches that it samples that happen to be near the center of the image.

\begin{table}[!t]
\centering
\caption{\textbf{Zero-shot classification performance on ImageNet-1K} when we pre-train the model on CC12M using the random, inverse Gaussian and Gaussian masking strategies to mask 50\% image patches.
The image encoder is ViT-B/16. We set $\sigma$ to 0.2.}
\begin{tabular}{l|c|c}
\hline

     \multirow{2}{*}{Method} & \multicolumn{2}{c}{0-shot} \\ 
        &   Before tuning & After tuning   \\
        \cline{1-3}
            random masking &  34.0 & 34.2 \\
            Inverse Gaussian Masking   &  31.1 & 30.9 \\ 
            Gaussian masking &  \textbf{35.1} & \textbf{35.4} \\ \hline
\end{tabular}
\label{table:rec}
\end{table}


\begin{figure}[!t]
    \centering
    \includegraphics[width=0.8\linewidth]{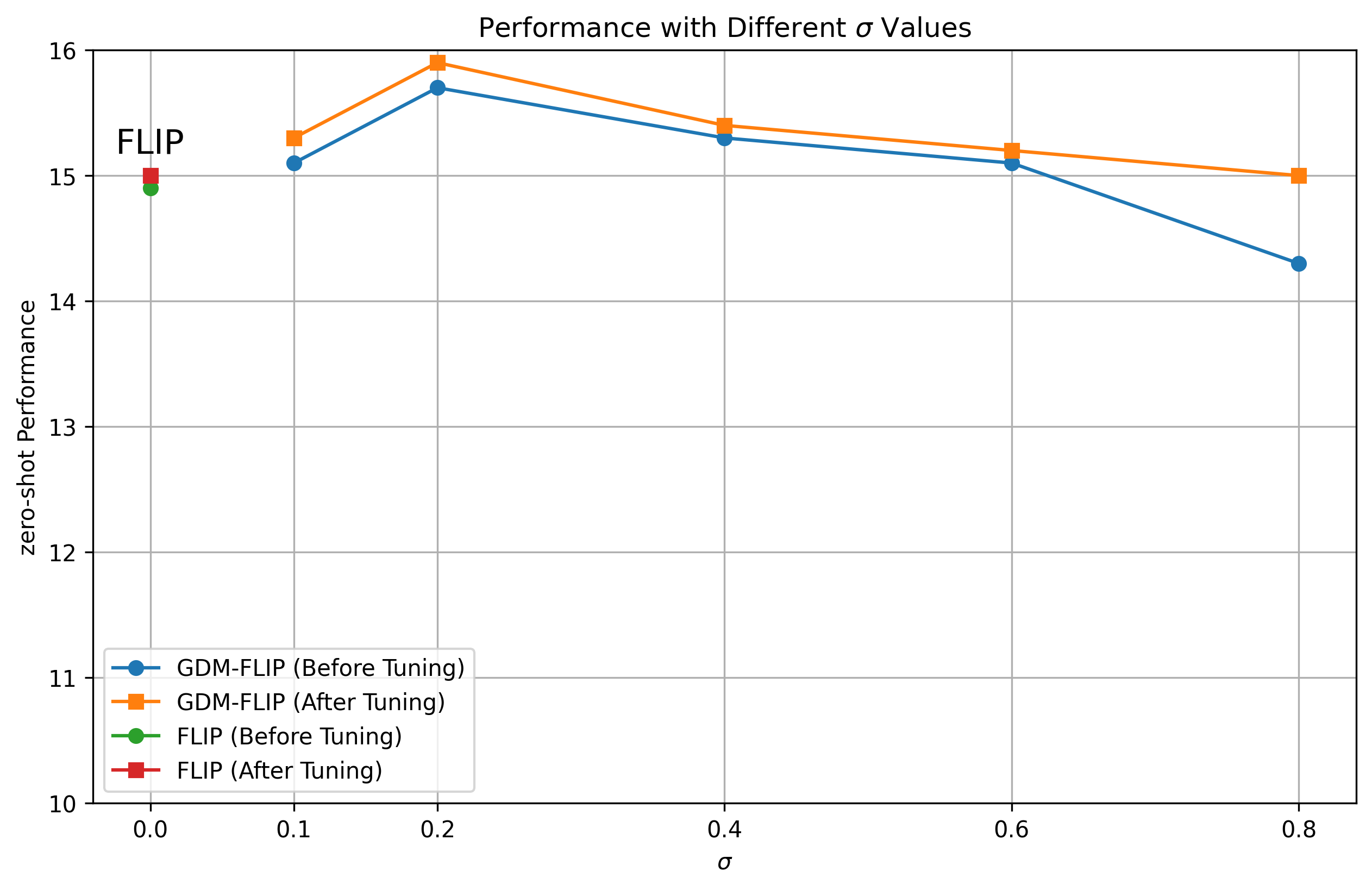}
    \caption{\textbf{Zero-shot classification performance on ImageNet-1K with different values of $\sigma$ (cf. Formula~\ref{fig:random_gaussian_mask}).}
    The models, using ViT-B/16 as the image encoder, were pre-trained on the CC3M dataset for 30 epochs and fine-tuned for one more epoch without masking.}
    \label{fig:sigma}
\end{figure}

\textbf{Different $\sigma$} 
For our experiments, the value of $\sigma$ was initially set by visualizations such as those in Figure~\ref{fig:exam} and with limited exploratory experiments. 
Here, we investigate the sensitivity of the model to $\sigma$ 
We again study a standard task, i.e., ImageNet-1K zero-shot, and train models with different \(\sigma\) values on the CC3M dataset. 
Results are presented in Figure~\ref{fig:sigma}.
From 0.1 to 0.2, the performance of the model improves by 0.6\%, and the model achieves optimal performance at \(\sigma = 0.20\), which is the value that we had chosen via visualisation. 
An increase in $\sigma$ from 0.2 results in decreased performance. 
However, the model still surpasses FLIP, which utilizes a random masking strategy, for $\sigma$ values up to 0.8 after fine-tuning. 
Thus, it is essential to balance the model's focus on both the image center and its edges, ensuring comprehensive attention across the entire image.
Further, from this experiment, we conclude that GLIP has a certain robustness to the choice of $\sigma$, since a $\sigma$ somewhat larger than the optimal size will decrease performance only mildly and still allow GLIP to outperform FLIP. 


\section{Conclusion and Outlook}

In this paper, we have introduced GLIP, a novel image masking strategy that 
improves the pre-training efficiency of Vision-Language Models. 
Our method outperforms the FLIP random image masking strategy by increasing the retention of image patches at the center of images. 
We achieved better performance than FLIP in various zero-shot tasks such as classification and image text retrieval.
Unlike A-CLIP, our approach simplifies the process by eliminating the need for additional inference to compute the attention scores of image patches.
Across a wide range of tasks and datasets, GLIP demonstrates superior performance compared to its FLIP counterparts, particularly with the use of a large masking ratio during pre-training. 
We also demonstrated that the performance of GLIP is similar to that of CLIP when we pre-train the model on the CC12M dataset.

Moving forward, it will be interesting to study the improvement offered by GLIP on very large-scale datasets.
In the original work on FLIP~\cite{li2023flip}, experiments were reported on training sets with 400 million samples and 2 billion samples.
Because FLIP is simply sampling patches, there are no inherent constraints that prevent it from scaling to larger data.
We point out that GLIP, for the same reason, also lacks any inherent constraints. 
For this reason, we are confident that GLIP will also deliver performance improvement on datasets much larger than the set of 10 million training examples we used here. 
Further, because GLIP is effective with a large masking ratio, it can process more samples than FLIP given the same training time. 
For this reason, we expect that GLIP will continue to show stronger competitiveness as datasets grow larger.

An interesting aspect of GLIP is its applicability across a wide range of datasets, as demonstrated by our experiments.
Intuitively, it might be expected that GLIP performs better on datasets with center focus, i.e., they contain images that were taken by a photographer who was actively framing the images so that the main subject material would be close to the center of the image, and not at the edge.
However, we have seen that GLIP shows a strong advantage over FLIP on data where there is apparently no explicit center focus, specifically, for the EuroSAT dataset~\cite{helber2019eurosat}, which contains remote sensing images, and PCam~\cite{ehteshami2017pcam,Veeling2018PCam}, which contains medical images of tumors. 
It remains an open question as to why GLIP performs so well on datasets where the main information does not have a strong tendency to be located towards the middle of the image.
A possible explanation is that patches at the center of the image are simply more helpful to the vision encoder because they are surrounded by full context.
A patch located at the periphery of an image does not enjoy a full context because the image ends at the image edge. 
Moving forward, we will continue to explore the nature of the reasons for which the centered masking of GLIP delivers an improvement over the random masking of FLIP.

\bibliographystyle{unsrt}  
\bibliography{references}

\begin{thebibliography}{10}

\bibitem{CLIP2021radford}
Alec Radford, Jong~Wook Kim, Chris Hallacy, Aditya Ramesh, Gabriel Goh, Sandhini Agarwal, Girish Sastry, Amanda Askell, Pamela Mishkin, Jack Clark, Gretchen Krueger, and Ilya Sutskever.
\newblock Learning transferable visual models from natural language supervision.
\newblock In {\em International Conference on Machine Learning}, 2021.

\bibitem{li2023flip}
Yanghao Li, Haoqi Fan, Ronghang Hu, Christoph Feichtenhofer, and Kaiming He.
\newblock Scaling language-image pre-training via masking.
\newblock In {\em IEEE/CVF Conference on Computer Vision and Pattern Recognition}, 2023.

\bibitem{he2022mae}
Kaiming He, Xinlei Chen, Saining Xie, Yanghao Li, Piotr Doll{\'a}r, and Ross Girshick.
\newblock Masked autoencoders are scalable vision learners.
\newblock In {\em IEEE/CVF conference on computer vision and pattern recognition}, pages 16000--16009, 2022.

\bibitem{He2020Momentum}
Kaiming He, Haoqi Fan, Yuxin Wu, Saining Xie, and Ross Girshick.
\newblock Momentum contrast for unsupervised visual representation learning.
\newblock In {\em 2020 IEEE/CVF Conference on Computer Vision and Pattern Recognition}, 2020.

\bibitem{deng2009imagenet}
Jia Deng, Wei Dong, Richard Socher, Li-Jia Li, Kai Li, and Li~Fei-Fei.
\newblock {Imagenet:} {A large-scale hierarchical image database}.
\newblock In {\em IEEE Conference on Computer Vision and Pattern Recognition}, 2009.

\bibitem{changpinyo2021cc12m}
Soravit Changpinyo, Piyush Sharma, Nan Ding, and Radu Soricut.
\newblock {Conceptual 12M}: Pushing web-scale image-text pre-training to recognize long-tail visual concepts.
\newblock In {\em IEEE/CVF Computer Vision and Pattern Recognition Conference}, 2021.

\bibitem{yang2023aclip}
Yifan Yang, Weiquan Huang, Yixuan Wei, Houwen Peng, Xinyang Jiang, Huiqiang Jiang, Fangyun Wei, Yin Wang, Han Hu, Lili Qiu, et~al.
\newblock Attentive mask clip.
\newblock In {\em IEEE/CVF International Conference on Computer Vision}, pages 2771--2781, 2023.

\bibitem{scaling2021Jia}
Chao Jia, Yinfei Yang, Ye~Xia, Yi-Ting Chen, Zarana Parekh, Hieu Pham, Quoc Le, Yun-Hsuan Sung, Zhen Li, and Tom Duerig.
\newblock Scaling up visual and vision-language representation learning with noisy text supervision.
\newblock In {\em International Conference on Machine Learning}, 2021.

\bibitem{bahdanau2014neural}
Dzmitry Bahdanau, Kyunghyun Cho, and Yoshua Bengio.
\newblock Neural machine translation by jointly learning to align and translate.
\newblock In {\em International Conference on Learning Representations, 2015}, 2015.

\bibitem{Jan2015NIPS}
Jan~K Chorowski, Dzmitry Bahdanau, Dmitriy Serdyuk, Kyunghyun Cho, and Yoshua Bengio.
\newblock Attention-based models for speech recognition.
\newblock In {\em Advances in Neural Information Processing Systems}, volume~28. Curran Associates, Inc., 2015.

\bibitem{lu2016hierarchical}
Jiasen Lu, Jianwei Yang, Dhruv Batra, and Devi Parikh.
\newblock Hierarchical question-image co-attention for visual question answering.
\newblock {\em Advances in neural information processing systems}, 29, 2016.

\bibitem{mnih2014recurrent}
Volodymyr Mnih, Nicolas Heess, Alex Graves, et~al.
\newblock Recurrent models of visual attention.
\newblock {\em Advances in neural information processing systems}, 27, 2014.

\bibitem{NIU202148}
Zhaoyang Niu, Guoqiang Zhong, and Hui Yu.
\newblock A review on the attention mechanism of deep learning.
\newblock {\em Neurocomputing}, 452:48--62, 2021.

\bibitem{NEURIPS2020Cubuk}
Ekin~Dogus Cubuk, Barret Zoph, Jon Shlens, and Quoc Le.
\newblock Randaugment: Practical automated data augmentation with a reduced search space.
\newblock In H.~Larochelle, M.~Ranzato, R.~Hadsell, M.F. Balcan, and H.~Lin, editors, {\em Advances in Neural Information Processing Systems}, volume~33, pages 18613--18624, 2020.

\bibitem{devries2017dataset}
Terrance DeVries and Graham~W Taylor.
\newblock Dataset augmentation in feature space.
\newblock {\em arXiv preprint arXiv:1702.05538}, 2017.

\bibitem{krizhevsky2012alexnet}
Alex Krizhevsky, Ilya Sutskever, and Geoffrey~E Hinton.
\newblock Imagenet classification with deep convolutional neural networks.
\newblock {\em Advances in neural information processing systems}, 25, 2012.

\bibitem{shorten2019survey}
Connor Shorten and Taghi~M Khoshgoftaar.
\newblock A survey on image data augmentation for deep learning.
\newblock {\em Journal of big data}, 6(1):1--48, 2019.

\bibitem{schuhmann2022laionb}
Christoph Schuhmann, Romain Beaumont, Richard Vencu, Cade~W Gordon, Ross Wightman, Mehdi Cherti, Theo Coombes, Aarush Katta, Clayton Mullis, Mitchell Wortsman, Patrick Schramowski, Srivatsa~R Kundurthy, Katherine Crowson, Ludwig Schmidt, Robert Kaczmarczyk, and Jenia Jitsev.
\newblock {LAION}-{5B}: An open large-scale dataset for training next generation image-text models.
\newblock In {\em Conference on Neural Information Processing Systems, Datasets and Benchmarks Track}, 2022.

\bibitem{Schuhmann2021laion400m}
Christoph Schuhmann, Richard Vencu, Romain Beaumont, Robert Kaczmarczyk, Clayton Mullis, Aarush Katta, Theo Coombes, Jenia Jitsev, and Aran Komatsuzaki.
\newblock {LAION-400M}: Open dataset of {CLIP-Filtered} 400 million image-text pairs, 2021.

\bibitem{sharma2018cc3m}
Piyush Sharma, Nan Ding, Sebastian Goodman, and Radu Soricut.
\newblock {Conceptual Captions:} a cleaned, hypernymed, image alt-text dataset for automatic image captioning.
\newblock In {\em Annual Meeting of the Association for Computational Linguistics}, 2018.

\bibitem{Dong2023MaskCLIP}
Xiaoyi Dong, Jianmin Bao, Yinglin Zheng, Ting Zhang, Dongdong Chen, Hao Yang, Ming Zeng, Weiming Zhang, Lu~Yuan, Dong Chen, Fang Wen, and Nenghai Yu.
\newblock {MaskCLIP:} masked self-distillation advances contrastive language-image pretraining.
\newblock In {\em IEEE/CVF Conference on Computer Vision and Pattern Recognition}, pages 10995--11005, 2023.

\bibitem{assran2022msn}
Mahmoud Assran, Mathilde Caron, Ishan Misra, Piotr Bojanowski, Florian Bordes, Pascal Vincent, Armand Joulin, Mike Rabbat, and Nicolas Ballas.
\newblock Masked siamese networks for label-efficient learning.
\newblock In {\em European Conference on Computer Vision}, pages 456--473, 2022.

\bibitem{dosovitskiy2020ViT}
Alexey Dosovitskiy, Lucas Beyer, Alexander Kolesnikov, Dirk Weissenborn, Xiaohua Zhai, Thomas Unterthiner, Mostafa Dehghani, Matthias Minderer, Georg Heigold, Sylvain Gelly, Jakob Uszkoreit, and Neil Houlsby.
\newblock An image is worth 16x16 words: Transformers for image recognition at scale.
\newblock In {\em International Conference on Learning Representations}, 2021.

\bibitem{arnheim1954art}
Rudolf Arnheim.
\newblock {\em Art and visual perception: A psychology of the creative eye}.
\newblock Univ of California Press, 1954.

\bibitem{roy1979art}
Roy~H. Quan.
\newblock Photography and the creation of meaning.
\newblock {\em Art Education}, 32(2):4--9, 1979.

\bibitem{tatler2007central}
Benjamin~W Tatler.
\newblock The central fixation bias in scene viewing: Selecting an optimal viewing position independently of motor biases and image feature distributions.
\newblock {\em Journal of vision}, 7(14):4--4, 2007.

\bibitem{chen2020simple}
Ting Chen, Simon Kornblith, Mohammad Norouzi, and Geoffrey Hinton.
\newblock A simple framework for contrastive learning of visual representations.
\newblock In {\em International Conference on Machine Learning}, 2020.

\bibitem{oord2018InfoNCE}
Aaron van~den Oord, Yazhe Li, and Oriol Vinyals.
\newblock Representation learning with contrastive predictive coding, 2019.

\bibitem{Gabriel2021openclip}
Gabriel Ilharco, Mitchell Wortsman, Ross Wightman, Cade Gordon, Nicholas Carlini, Rohan Taori, Achal Dave, Vaishaal Shankar, Hongseok Namkoong, John Miller, Hannaneh Hajishirzi, Ali Farhadi, and Ludwig Schmidt.
\newblock {OpenCLIP}, 2021.

\bibitem{li2022blip}
Junnan Li, Dongxu Li, Caiming Xiong, and Steven Hoi.
\newblock {BLIP:} bootstrapping language-image pre-training for unified vision-language understanding and generation.
\newblock In {\em ICML}, 2022.

\bibitem{mu2021slip}
Norman Mu, Alexander Kirillov, David~A. Wagner, and Saining Xie.
\newblock {SLIP:} {Self-supervision Meets Language-Image Pre-training}.
\newblock In {\em European conference on computer vision}, 2022.

\bibitem{Ashish2017Transformer}
Ashish Vaswani, Noam Shazeer, Niki Parmar, Jakob Uszkoreit, Llion Jones, Aidan~N. Gomez, Lukasz Kaiser, and Illia Polosukhin.
\newblock Attention is all you need.
\newblock In {\em Advances in Neural Information Processing Systems}, 2017.

\bibitem{lin2014coco}
Tsung-Yi Lin, Michael Maire, Serge Belongie, James Hays, Pietro Perona, Deva Ramanan, Piotr Doll{\'a}r, and C~Lawrence Zitnick.
\newblock {Microsoft COCO:} {Common objects in context}.
\newblock In {\em European Conference on Computer Vision}, pages 740--755, 2014.

\bibitem{young2014Flickr30k}
Peter Young, Alice Lai, Micah Hodosh, and Julia Hockenmaier.
\newblock From image descriptions to visual denotations: New similarity metrics for semantic inference over event descriptions.
\newblock {\em Transactions of the Association for Computational Linguistics}, pages 67--78, 2014.

\bibitem{helber2019eurosat}
Patrick Helber, Benjamin Bischke, Andreas Dengel, and Damian Borth.
\newblock Eurosat: A novel dataset and deep learning benchmark for land use and land cover classification.
\newblock {\em IEEE Journal of Selected Topics in Applied Earth Observations and Remote Sensing}, 12(7):2217--2226, 2019.

\bibitem{ehteshami2017pcam}
Babak Ehteshami~Bejnordi, Mitko Veta, Paul Johannes~van Diest, Bram van Ginneken, Nico Karssemeijer, Geert Litjens, Jeroen A. W.~M. van~der Laak, , and the CAMELYON16~Consortium.
\newblock {Diagnostic Assessment of Deep Learning Algorithms for Detection of Lymph Node Metastases in Women With Breast Cancer}.
\newblock {\em JAMA}, 318(22):2199--2210, 2017.

\bibitem{Veeling2018PCam}
Bastiaan~S. Veeling, Jasper Linmans, Jim Winkens, Taco Cohen, and Max Welling.
\newblock Rotation equivariant cnns for digital pathology.
\newblock In {\em Medical Image Computing and Computer Assisted Intervention}, 2018.

\bibitem{adam11stl10}
Adam Coates, Andrew Ng, and Honglak Lee.
\newblock An analysis of single-layer networks in unsupervised feature learning.
\newblock In {\em International Conference on Artificial Intelligence and Statistics}, 2011.

\bibitem{krizhevsky2009cifar}
Alex Krizhevsky, Geoffrey Hinton, et~al.
\newblock Learning multiple layers of features from tiny images.
\newblock 2009.

\bibitem{lecun1998mnist}
Yann LeCun, L{\'e}on Bottou, Yoshua Bengio, and Patrick Haffner.
\newblock Gradient-based learning applied to document recognition.
\newblock {\em Proceedings of the IEEE}, 86(11):2278--2324, 1998.

\end{thebibliography}

\end{document}